\documentclass[11pt,a4paper]{article}
\usepackage[hyperref]{acl2021}
\usepackage{times}
\usepackage{latexsym}

\usepackage{microtype}
\usepackage{amsfonts}
\usepackage{amsmath}
\usepackage{amssymb}
\usepackage{breqn}
\usepackage{graphicx}
\usepackage{tikz}
\usepackage{xspace}
\usepackage{xcolor}
\usepackage[shortlabels]{enumitem}
\usepackage{arydshln}
\usepackage{mathtools}
\usepackage{subcaption}

\aclfinalcopy % Uncomment this line for the final submission
\def\aclpaperid{1585} %  Enter the acl Paper ID here

\newcommand{\change}[1]{{#1}}

\newtheorem{theorem}{Theorem}[section]

\newtheorem{innercustomthm}{Theorem}
\newenvironment{customthm}[1]
  {\renewcommand\theinnercustomthm{#1}\innercustomthm}
  {\endinnercustomthm}

\newcommand{\pd}[2]{\frac{\partial #1}{\partial #2}}
\newcommand{\pdi}[2]{{\partial #1}/{\partial #2}}

\newcommand{\normeq}{\stackrel{\mathclap{\mbox{\tiny$\Theta$}}}{=}}
\newcommand{\ptheta}{\pmb \theta}
\newcommand{\px}{\pmb x}  
\newcommand{\prv}{\pmb r^v}
\newcommand{\rv}{r^v}
\newcommand{\prk}{\pmb r^k}
\newcommand{\norm}[1]{\lVert#1\rVert}

\title{Optimizing Deeper Transformers on Small Datasets}

\author{Peng Xu$^1$,
        Dhruv Kumar$^{*1,2}$,
        Wei Yang$^1$,
        Wenjie Zi$^1$,
        Keyi Tang$^1$,
        Chenyang Huang\thanks{\ Work done while the author was an intern in Borealis AI.} $^{1,5}$, \\
        \bf Jackie Chi Kit Cheung$^{1,3,4}$, Simon J.D. Prince$^1$,
        Yanshuai Cao$^1$\\
  $^1$Borealis AI $^2$University of Waterloo \\ $^3$McGill University $^4$Canada CIFAR Chair, Mila $^5$University of Alberta  \\
  \scriptsize \texttt{\{peng.z.xu, wei.yang, wenjie.zi, keyi.tang, simon.prince, yanshuai.cao\}@borealisai.com} \\
  \scriptsize \texttt{dhruv.kumar@uwaterloo.ca}, \texttt{chuang8@ualberta.ca}, \texttt{jcheung@cs.mcgill.ca} \\
  }

\date{}

\begin{document}
\maketitle
\begin{abstract}

It is a common belief that training deep transformers from scratch requires large datasets. Consequently, for small datasets, people usually use shallow and simple additional layers on top of pre-trained models during fine-tuning.
This work shows that this does not always need to be the case: with proper initialization and optimization, the benefits of very deep transformers can carry over to challenging tasks with small datasets, including Text-to-SQL semantic parsing and logical reading comprehension.
In particular, we successfully train $48$ layers of transformers, comprising $24$ fine-tuned layers from pre-trained RoBERTa and $24$ relation-aware layers trained from scratch.
With fewer training steps and no task-specific pre-training, we obtain the state-of-the-art performance on the challenging cross-domain Text-to-SQL parsing benchmark Spider\footnote{The code to reproduce our results can be found in: \href{https://github.com/BorealisAI/DT-Fixup}{\color{blue} https://github.com/BorealisAI/DT-Fixup}}.
We achieve this by deriving a novel \textbf{D}ata-dependent \textbf{T}ransformer \textbf{Fix}ed-\textbf{up}date initialization scheme (DT-Fixup), inspired by the prior T-Fixup work \cite{huang2020improving}.
Further error analysis shows that increasing depth can help improve generalization on small datasets for hard cases that require reasoning and structural understanding.

%Due to the common belief that training deep transformers from scratch requires large datasets, people usually only use shallow and simple additional layers on top of pre-trained models during fine-tuning on small datasets.
%We provide evidence that this does not always need to be the case: with proper initialization and training techniques, the benefits of very deep transformers are shown to carry over to hard structural prediction tasks,  even using small datasets.
%In particular, we successfully train $48$ layers of transformers for a semantic parsing task. These comprise $24$ fine-tuned transformer layers from pre-trained RoBERTa and $24$ relation-aware transformer layers trained from scratch.
%With fewer training steps and no task-specific pre-training, we obtain the state of the art performance on the challenging cross-domain Text-to-SQL semantic parsing benchmark Spider.
%We achieve this by deriving a novel \textbf{D}ata dependent \textbf{T}ransformer \textbf{Fix}ed-\textbf{up}date initialization scheme (DT-Fixup), inspired by the prior T-Fixup work \cite{huang2020improving}.
%Further error analysis demonstrates that increasing the depth of the transformer model can help improve generalization on the cases requiring reasoning and structural understanding\footnote{Work in progress.}.

\end{abstract}

\section{Introduction}
In recent years, large-scale pre-trained language models \cite{radford2019language,devlin2018bert, liu2019roberta} trained with transformers \cite{vaswani2017attention} have become standard building blocks of modern NLP systems to help improve generalization when task-specific annotations are limited. 
%Most NLP tasks employ shallow and simple additional neural components on top of the pre-trained representations, such as a classifier head.
In practice, it has been found that deeper transformers generally yield better results with sufficient training data \cite{lan2019albert}, especially on tasks involving reasoning and structural understanding.
This suggests that additional transformer layers should be employed in conjunction with pre-trained models, instead of simple and shallow neural components, such as a classifier head, currently used by models of many NLP tasks.
However, the common belief in the literature is that training deep transformers from scratch requires large datasets, and few attempts have been made on small datasets, to the best of our knowledge.
One implication is that although extra transformer layers on top of pre-trained models should help with more challenging problems in principle, it does not work in practice due to limited training data.
We show that after resolving several optimization issues with the method proposed in this work, it is possible to train very deep transformers with improved generalization even on small datasets.

% 

%However with limited task-specific annotated data, fine-tuning large-scale pre-trained models have become a common practice in many NLP applications that aim to improve generalization ability.
%In such use cases, a mixed setup of additional yet-to-be-trained transformers on top of pre-trained models is required.
One advantage of pre-trained models is the reduced computational resources needed when fine-tuning on small datasets.
For instance, it allows practitioners to finetune on a single GPU and obtain strong performance on a downstream task. 
However, the large size of pre-trained models limits the batch size that can be used in training new transformer layers on a small computational budget. 
Despite their broad applications, training transformer models is known to be difficult \citep{popel2018training}.
The standard transformer training approach leverages learning rate warm-up, layer normalization \cite{ba2016layer} and a large batch size, and models typically fail to learn when missing any one of these components.
The restricted batch size aggravates the training difficulties.
Even if a large batch size can be feasibly employed, poorer generalization results are often observed \cite{keskar2016large}, especially when the dataset size is only several times larger than the batch size.
Furthermore, many recent works noticed a performance gap in this training approach due to layer normalization \cite{xu2019lipschitz, nguyen2019transformers, zhang2019improving, wang2019learning, liu2020understanding, huang2020improving}. 

Inspired by the recent T-Fixup by \citet{huang2020improving}, which eliminates the need for learning rate warm-up and layer normalization to train vanilla transformers, we derive a data-dependent initialization strategy by applying different analyses to address several key limitations of T-Fixup.
We call our method the \textbf{D}ata-dependent \textbf{T}ransformer \textbf{Fix}ed-\textbf{up}date initialization scheme, {\em DT-Fixup}.
In the mixed setup of additional yet-to-be-trained transformers on top of pre-trained models, DT-Fixup enables the training of significantly deeper transformers, and is generally applicable to different neural architectures.
Our derivation also extends beyond vanilla transformers to transformers with relational encodings \cite{shaw2018self}, allowing us to apply the results to one variant called relation-aware transformer \cite{{wang2019rat}}.
By applying DT-Fixup on different tasks, we show that the impression that deep transformers do not work on small datasets stems from the optimization procedure rather than the architecture.
With proper initialization and optimization, training extra transformer layers is shown to facilitate the learning of complex relations and structures in the data.

We verify the effectiveness of DT-Fixup on Spider \cite{yu2018spider}, a complex and cross-domain Text-to-SQL semantic parsing benchmark, and ReColr \cite{yu2020reclor}, a reading comprehension dataset requiring logical reasoning.
While Text-to-SQL semantic parsing is inherently different from reading comprehension, they share similar characteristics which require certain levels of reasoning and structural understanding ability.
Meanwhile, the sizes of both datasets are less than 10k training samples, which is tiny by deep learning standards \change{and renders large-batch training undesirable due to poor generalization\footnote{For a comparison, T-Fixup applies batch sizes of more than 1k on machine translation to stabilize the training, which would hurt the generalization significantly on our datasets whose sizes are less than 10k.}}.
% The correct predictions depend on the interplay between the questions and the schema structures and the generalization over unseen schemas during inference.
% As a result, reasoning and structural understanding are crucial to perform well on this task, especially for the more challenging cases.
% Meanwhile, the dataset size is tiny by deep learning standards, with only $10,181$ questions and $5,693$ queries covering $200$ databases in $138$ domains.

On both datasets, DT-Fixup consistently outperforms the standard approach with better generalization and allows the training of significantly deeper transformer models.
For Spider, we successfully apply DT-Fixup to train a Text-to-SQL parser containing $48$ transformer layers, with $24$ relation-aware layers trained from scratch on top of $24$ pre-trained layers from pre-trained RoBERTa \cite{liu2019roberta}.
Our parser achieves $70.9\%$ exact match accuracy on the Spider test set, which is the state of the art at the time of writing.
At the same time, it requires less training steps and no task-specific pre-training as compared to the prior art \cite{yu2020grappa}.
For ReClor, we rank the second on the public leaderboard by simply adding $4$ transformer layers on top of RoBERTa.
Further error analysis shows that the performance improvements by increasing the depth mainly come from better generalization on the harder cases requiring reasoning and structural understanding.
Even the failed predictions from the deep models are more reasonable than from the shallow ones.

\section{Background}
In this section, we present the necessary background by first introducing the relation-aware transformer layer, which outperforms the vanilla transformer layer with limited data by injecting additional inductive bias \cite{wang2019rat}.
Then, we introduce the T-Fixup technique \cite{huang2020improving} for optimizing deeper vanilla transformers and discuss why it does not directly apply in the mixed transformer optimization setup.

% \vspace{-0.5em}

\subsection{Relative Position and Relational Encodings in Transformers}
\label{sec:rel_transformer}

Consider a set of inputs $X=[\pmb{x}_1, \dots, \pmb{x}_n]$ where $\pmb{x}_i\in\mathbb{R}^{d_x}$.
% In general, we consider it an unordered set, although $\pmb{x}_i$ may be imbued with positional embeddings to add an explicit ordering relation.
A \emph{transformer}, introduced by \citet{vaswani2017attention}, is a stack of blocks, with each block consisting of a multi-head \emph{self-attention layer}, layer normalizations, a multi-layer perceptron and skip connections. Each block (with one head in self-attention for notational simplicity) transforms each $\pmb{x}_i$ into $\pmb{y}_i\in\mathbb{R}^{d_x}$ as follows:
\begin{align}
        \alpha_{ij} & =\text{softmax}\left({\pmb{x}_i\pmb q(\pmb{x}_j\pmb k)^\top} \middle/ {\sqrt{d_z}}\right) \label{eq:trans1} \\
        \pmb{z}_i &= {\textstyle\sum}_{j=1}^n \alpha_{ij}\pmb{x}_j \pmb v;  \label{eq:trans2}\\
       \pmb{\tilde y}_i & = \text{LayerNorm}(\pmb{x}_i + \pmb{z}_i\pmb w^\top)  \label{eq:trans3}\\
        \pmb{y}_i & = \text{LayerNorm}(\pmb{\tilde y}_i + \text{MLP}(\pmb{\tilde y}_i)) \label{eq:trans4}
\end{align}
\noindent where the $\text{softmax}$ operation is applied across the index $j$, $\text{MLP}$ is a two-layer perceptron, LayerNorm is a \emph{layer normalization} \citep{ba2016layer} layer, and $\pmb q, \pmb k, \pmb v\in\mathbb{R}^{d_x \times d_z}, \pmb w \in\mathbb{R}^{d_x \times d_z}$.

In order to bias the transformer toward some pre-existing relational features between the inputs, \citet{shaw2018self} described a way to represent \emph{relative position information} in a self-attention layer by changing Equation~\ref{eq:trans1}-\ref{eq:trans2} as follows:
\begin{equation}\label{eq:relational}
    \begin{split}
        \alpha_{ij} & =\text{softmax}\left(\frac{\pmb{x}_i\pmb q(\pmb{x}_j\pmb k + {\color{red}\pmb{r}_{ij}^k})^\top}{\sqrt{d_z}}\right) \\
        \pmb{z}_i & = {\textstyle\sum}_{j=1}^n \alpha_{ij}(\pmb{x}_j \pmb v + {\color{red} \pmb{r}_{ij}^v})
    \end{split}
\end{equation}
\noindent Here the $\pmb{r}_{ij}\in\mathbb{R}^{d_z}$ terms encode the known relationship between two elements $\pmb x_i$ and $\pmb x_j$ in the input.
\citet{wang2019rat} adapted this framework to effectively encode the schema information using $\pmb{r}_{ij}$'s for Text-to-SQL parsers, and called it relation-aware transformer (RAT).

% \vspace{-0.5em}

\subsection{T-Fixup and its Limitations}\label{sec:tfixup}
\citet{huang2020improving} found that the requirement for the warmup during the early stage training of the transformers comes from a combined effect of high variance in the Adam optimizer and back-propagation through layer normalization.
Bounding the gradient updates would reduce the variance and make training stable, which can be achieved by appropriately initializing the model weights.

They derived a weight initialization scheme called T-Fixup for the vanilla transformer that fully eliminates the need for layer normalization and learning rate warmup, and stabilizes the training to avoid harmful plateaus of poor generalization. 
T-Fixup requires the inputs $\pmb x$ to be Gaussian randomly initialized embeddings with variance $d^{-\frac{1}{2}}$ where $d$ is the embedding dimension.
Then, the input and parameters of the encoder, $\pmb x$, $\pmb v$, $\pmb w$ in the vanilla self-attention blocks as well as the weight matrices in the MLP blocks defined in Eq.~\ref{eq:trans1}-\ref{eq:trans4} are re-scaled by multiplying with a factor of $0.67N^{-\frac{1}{4}}$, where $N$ are the number of transformer layers.
% Yanshuai/Peng This last sentence does not make sense.  What exactly is being rescaled?

However, there are two restrictions of T-Fixup narrowing down the range of its application.
First, T-Fixup is only designed for vanilla transformer but not other variants like the relative position or relation-aware version described previously.
Second, they make the critical assumption that the inputs $\pmb x$ can be freely initialized then scaled to the same magnitude as $\pmb v$, $\pmb w$ and MLP weights. 
This renders the method inapplicable for the mixed setup where the inputs to the yet-to-be-trained transformer layers depend on the outputs from the pre-trained models.
The first issue can be addressed by re-deriving the scaling factor following the methodology of T-Fixup but taking into account the additional relational term.
However, to lift the second restriction requires changing the assumption and more dramatic modification to the analysis.
\begin{figure}[h]
	\begin{center}
	\includegraphics[width=0.85\linewidth]{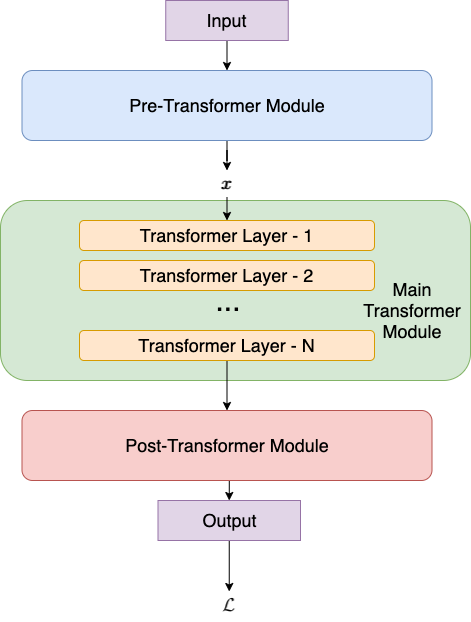}
	\caption{Illustration of the general neural architecture on which our method can be applied.}
	\label{fig:brto}
	\end{center}
\end{figure}
% This figure has layers G1...GN but in the text of section 4.1 there is G1...G2N

% \vspace{-0.5em}

\section{Our Approach}
\label{sec:main_method}
We now follow the analysis framework of T-Fixup \cite{huang2020improving}, but derive the conditions to bound the gradient updates of the self-attention block in the presence of a pre-trained model.
Based on the derivation, we propose a data-dependent initialization strategy for the mixed setup of the new transformers on pre-trained encodings.

%They proposed a weight initialization schema for the vanilla transformer that eliminates the need for layer normalization and warmup.
% However, their method is only applicable on vanilla transformer and depends on the assumption that the scale of the inputs to the transformer can be controlled through initialization, which is not always true in practice.

% \vspace{-0.5em}

\subsection{Applicable Architectures}\label{sec:aa}
Our analysis applies to the general architecture type illustrated in Figure~\ref{fig:brto}, where the input passes through a pre-transformer, a main transformer, and a post-transformer module before outputting. The pre and post transformer modules can be any architectures that can be stably trained with Adam \cite{kingma2014adam}, including MLP, LSTM, CNN, or a pre-trained deep transformer module which can be stably fine-tuned with a learning rate significantly smaller than the main learning rate used for the main transformer module. 
For this work, we will just consider the case of the main transformer containing only the encoder for simplicity, while our decoder will be an LSTM which can be viewed as part of the post-transformer module.
Extending our analysis to include deep transformer decoder is straightforward following the framework of \citet{huang2020improving}.

%Yanshuai/Peng Second sentence.  How could the post-transformer module be a pre-trained transformer?  I don't think this makes sense and so it needs a bit of reworking.

We use $f_e$ to denote the pre-transformer module ($e$ for pre-trained encoder), and its parameters $\pmb\theta_e$; similarly  $f_o$ for post-transformer module ($o$ for output) with parameters $\pmb\theta_o$.
The main transformer module $f_G$ is a stack of $L$ transformer blocks, each consisting of a self-attention block and a MLP block.
Let $G_l, l=1,\dots,2N$ denote individual self-attention or MLP layers in the blocks ($G_l$'s do not include the skip connections), with parameters $\pmb\theta_l$ and let $L=2N$, $f_G$'s parameters are denoted by $\pmb\theta_G=\bigcup\limits_{l=1}^L \pmb\theta_l$.

\subsection{Theoretical Results for Stable Update}

Let the whole model with the output softmax layer(s) and all layer normalization blocks removed be denoted by $f(\cdot;\pmb \theta)$ and the loss function by $\mathcal{L}$, where $\pmb\theta$ are all the learnable parameters. Following \citet{huang2020improving}, we aim to derive a condition under which, per each SGD update with learning rate $\eta$, the model output changes by $\Theta(\eta)$, i.e. $\lVert \Delta f \rVert = \Theta(\eta)$ where $\Delta f = f(\cdot; \pmb \theta - \eta \pd{ \mathcal{L}}{\pmb\theta} ) - f(\cdot; \pmb \theta)$. By Taylor expansion, the SGD update is:
\begin{align}
    \Delta f = & \pd{ f}{\pmb\theta_o}\Delta \pmb\theta_o + \pd{ f}{ \pmb\theta_{G}}\Delta \pmb\theta_G + \pd{ f}{\pmb\theta_e}\Delta \pmb\theta_e + \nonumber \\
    \phantom{\Delta f =} & O(\lVert \pmb\theta_o\rVert^2 + \lVert \pmb \theta_{G} \rVert^2 + \lVert \pmb\theta_e \rVert^2) \nonumber \\
    \phantom{\Delta f} = & -\eta(\pd{ f_o}{ \pmb \theta_o}\pd{ f_o}{ \pmb \theta_o}^\top \pd{ \mathcal{L}}{ f_o}^\top + \nonumber \\
    \phantom{\Delta f =} & \pd{ f_o}{ f_G}\pd{ f_G}{ \pmb \theta_G}\pd{ f_G}{ \pmb \theta_G}^\top\pd{ f_o}{ f_G}^\top\pd{ \mathcal{L}}{ f_o}^\top + \nonumber \\
    \phantom{\Delta f =} & \pd{ f_o}{ f_G}\pd{ f_G}{ f_e}\pd{ f_e}{ \pmb \theta_e}\pd{ f_e}{ \pmb \theta_e}^\top\pd{ f_G}{ f_e}^\top\pd{ f_o}{ f_G}^\top\pd{ \mathcal{L}}{ f_o}^\top) \nonumber \\
    \phantom{\Delta f =} & +O(\eta^2)
\end{align}

As assumed in Sec.~\ref{sec:aa}, we can stably train $f_e$ and $f_o$ coupled with $\mathcal{L}$, \emph{i.e,} $\lVert \pd{ \mathcal L}{ f_o}\rVert= \lVert \pd{ f_o}{ \pmb\theta_o}\rVert=\lVert \pd{ f_e}{ \pmb\theta_e} \rVert = \lVert \pd{ f_o}{ f_G} \rVert = \lVert \pd{ f_G}{ f_e} \rVert = \Theta(1)$, we only need to bound the magnitudes of $\pd{ f_G}{ \pmb\theta_G} $ to bound the overall SGD update. %% which can be expanded as:
%% \begin{equation}
%%     \pd{ f_G}{\pmb\theta_G}\pd{ f_G}{\pmb\theta_G}^\top = \sum_{l=1}^L \pd{ G_l}{ \pmb \theta_{l}}\pd{ G_l}{ \pmb \theta_{l}}^\top + O(\sum_{l=1}^L \pd{ G_l}{ \pmb \theta_{l}}^4)
%% \end{equation}
%\noindent showing that bounding the magnitudes of gradient updates $\Delta G_l$ to be order of $\Theta(1/L)$ can make each SGD update bounded by $\Theta(\eta)$ per optimization step after initialization as $\eta \rightarrow 0$.
%Since each layer of the relational transformer has one attention block and one MLP block, $L=2N$.
% In other words, $\lVert \Delta f \rVert = \Theta(\eta)$, where $\Delta f \triangleq f(\cdot;\pmb\theta - \eta \pd{ \mathcal{L}}{ \pmb \theta}) - f(\cdot;\pmb\theta)$.
%\citet{zhang2019fixup} and \citet{huang2020improving} derived appropriate initialization to obtain $\Theta(1/L)$ bound for the MLP blocks and the vanilla self-attention blocks respectively.
% Empirically, the theoretical derivation for the SGD update also works well for Adam.
%We adapt their framework and derive a solution to the relation-aware self-attention blocks $G(\pmb x)=\pmb z \pmb w^\top$,
% \begin{equation}
%     % G(\pmb x) = \text{softmax}\left(\frac1{\sqrt{d_x}}\pmb x \pmb q(\pmb k \pmb x + \pmb r^k)^\top\right)(\pmb x \pmb v + \pmb r^v) \pmb w
%     G(\pmb x) = \pmb z \pmb w^\top
% \end{equation}
%where $\pmb z$ is defined in Eq.~\ref{eq:relational}.
Since what we care is the magnitude of the update as it relates to the depth, we can assume all parameters to be scalars, \emph{i.e,} $\pmb q_l, \pmb k_l, \pmb v_l, \pmb w_l, \pmb r^k_l, \pmb r^v_l$  reduce to scalars $q_l, k_l, v_l, w_l, r^k_l, r^v_l \in \mathbb{R}$. The next theorem states the condition under which, $\norm{\pd{f_G}{\ptheta_G}}$ is bounded by $\Theta(1)$, achieving the overall $\norm{\Delta f} = \Theta(\eta)$.
\begin{theorem} \label{main_theorem}
Assuming $\norm{\px} = \Theta(\mu)$ for some $\mu \gg 1$, then $\norm{\pd{f_G}{\ptheta_G}} = \Theta(1)$ if $\norm{v_l} = \norm{w_l} = \norm{\rv_l} = \Theta{\left(((4\mu^2 + 2\mu + 2)N)^{-\frac{1}{2}}\right)}$ for all encoder layers $l$ in relation-aware transformers; and $\norm{v_l} = \norm{w_l} = \Theta{\left((4\mu^2N)^{-\frac{1}{2}}\right)}$ in the case of vanilla transformers.
\end{theorem}
The proof is in Appendix \ref{sec:full_proof}. One important immediate observation is that our scaling as the depth $N$ is to the power of $-1/2$, whereas T-Fixup has a scaling with power of $-1/4$.

While this theorem is all we need for deriving our DT-Fixup approach, it is not immediately intuitive.
So next we inspect what it takes to bound the change in a individual layer output $\norm{\Delta G_l}$ to $\Theta(\eta/L)$ in each gradient update.
This will shine some light on the particular form of the expressions in Theorem \ref{main_theorem}:
\begin{theorem} \label{theorem:main}
Let $\pmb \px_l=[x^l_1, \dots, x^l_n]$ be the input into $l$-th layer, and assume that $\lVert  \partial \mathcal{L} /  \partial  G_l\rVert = \Theta(1)$, i.e.\ the gradient signal from the layers above is bounded, then $\Delta G_l = G_l(\pmb x_l - \eta\pd{ \mathcal{L}}{ \pmb x_l};\pmb \theta_l - \eta \pd{ \mathcal{L}}{ \mathcal{\pmb \theta}_l})-G_l(\pmb x_l;\pmb \theta_l)$ satisfies $\lVert \Delta G_l \rVert=\Theta(\eta/L)$ when for all $i=1,\dots, n$:
 \begin{equation}\label{eq:cond}
 \begin{split}
 &2\lVert v_l \rVert^2 \lVert x^l_i \rVert^2 + 2 \lVert v_l \rVert \lVert \rv_l \rVert \lVert x^l_i \rVert + \lVert \rv_l \rVert^2\\
 &+ \lVert w_l \rVert^2 (1 + 2 \lVert x^l_i \rVert^2) = \Theta(1/N) \\
 \end{split}
 \end{equation}  
 for relation-aware transformers.
 Alternatively, in the case of vannilla transformers:
 \begin{equation}\label{eq:cond2}
 \lVert v_l\rVert^2 \lVert x^l_i \rVert^2 + \lVert w_l \rVert^2 \lVert x^l_i \rVert^2 = \Theta(1/L)
 \end{equation}
\end{theorem}
In this case, the proof is straightforward by taking partial derivatives of $G_l$ with respect to each parameter, and keep the terms with the lowest powers as they dominate the norm when the scale is smaller than one. Appendix~\ref{sec:proof} gives the detailed proof.
The insight from this theorem is: if the input $\px_l$ has the same norm as $\px$, setting parameters $v_l, w_l, \rv_l$ to have the same norm and solve the equations would yield the scale factors in Theorem \ref{main_theorem}.

\paragraph{Remark:} 
In T-Fixup, the corresponding condition to Eq.~\ref{eq:cond2} keeps the term $\lVert v_l \rVert^2 \lVert w_l \rVert^2$ which is dropped by ours.
It is due to the fact that T-Fixup assumes $\lVert x_i \rVert$ can be controlled to be the same scale as $v_l$ and $w_l$, so the lowest power terms (which are dominating the norms here) are the quartic ($4$th power) ones.
For us, $\norm{\px}$ is treated separately by a constant to be estimated from data, so the lowest power terms are the quadratic ones in $v_l, w_l, \rv_l$ in Eq. \ref{eq:cond} and \ref{eq:cond2}, and $\lVert v_l \rVert^2 \lVert w_l \rVert^2$ are dropped.
Another important distinction from T-Fixup is that we assume the estimated $\norm{\px}$ to be much larger than the scale of $v_l$ and $w_l$, unlike the case when they are also controlled to be the same scale. 
As we will see next, these changes imply our proposed method employs more aggressive scaling for initialization as compared to T-Fixup, and the assumption that $\norm{\px}$ has larger scale is satisfied naturally.

\subsection{Proposed Method: DT-Fixup}
% Since each layer of the relational transformer has one attention block and one MLP block, $L=2N$ in Theorem~\ref{theorem:main}.

Unlike previous works \cite{zhang2019fixup, huang2020improving}, appropriate initialization is not enough to ensure Eq. \ref{eq:cond} and \ref{eq:cond2} during the early stage of the training.
This is due to the fact that the input $\pmb x$ often depends on the pre-trained model weights instead of being initialized by ourselves.
Empirically, we observe that the input norm $\lVert \pmb x \rVert$ are relatively stable throughout the training but difficulty to control directly by re-scaling.
Based on this observation, we treat $\lVert \pmb x \rVert$ as a constant and estimate it by a forward pass on all the training examples as $\mu=\max_j[\lVert \pmb x_j \rVert]$. We then use this estimated $\mu$ in the factors of Theorem \ref{main_theorem} to obtain the scaling needed for initialization. Since parameters of all layers are initialized to the same scale, we drop index $l$ for brevity in this section.
%% Assuming $\lVert v \rVert = \lVert w \rVert = \lVert r_i^v \rVert$,  equation~\ref{eq:cond} in Theorem~\ref{theorem:main} can be satisfied with $\lVert v \rVert = \lVert w \rVert = \lVert r_i^v \rVert = (N * (4\mu^2 + 2\mu + 2))^{-\frac12}$ for all $i=1,\dots,n$.
%% Assuming $\lVert v \rVert = \lVert w \rVert$, the equation~\ref{eq:cond2} in Theorem~\ref{theorem:main2} can be satisfied with $\lVert v \rVert = \lVert w \rVert = N^{-\frac12}/(2\mu)$.
In practice, $\mu$ is on the order of $10$ for pre-trained models, hence $v$, $w$ and $r_i^v$ are naturally two orders of magnitude smaller.
DT-Fixup is described as follows:

% Yanshuai/Peng. I'd like to see you generate this formula explicitly.

% In order to circumvent this issue, we add an additional learnable mapping $\pmb \pi$ at the end of the pre-transformer module $f_{\text{pre}}$, and initialize $\pmb \pi$ properly to make $\lVert \pmb x \rVert = \lVert \pmb\pi\pmb{\tilde x}\rVert = \Theta(1)$ instead, where $\pmb{\tilde x}$ is the original input without such a mapping.

\begin{itemize}
    \item Apply Xavier initialization \cite{glorot2010understanding} on all free parameters except loaded weights from the pre-training models;
    \item Remove the learning rate warm-up and all layer normalization in the transformer layers, except those in the pre-trained transformer;
    \item Forward-pass on all the training examples to get the max input norm $\mu=\max_j[\lVert \pmb{x}_j \rVert]$;
    \item Inside each transformer layer, scale $v, w, r^v$ in the attention block and weight matrices in the MLP block by $(N*(4\mu^2 + 2\mu + 2))^{-\frac12}$ for relation-aware transformer layer; or scale $v, w$ in the attention block and weight matrices in the MLP block by $N^{-\frac12}/(2\mu)$ for vanilla transformer layer.

\end{itemize}

% \yanshuai{Although the quadratic terms dominate, but that just applies to the fight between quartic and quadratic terms of the same variable; in the current derived fixup, the input embeddings are not scaled, so something is missing}

% \noindent In practice, this method can be applied on any task that requires a neural architecture similar to what is depicted in Figure~\ref{fig:brto}.
% Generally, $f_i$ and $f_o$ can be any standard neural architecture paired with any loss function $\mathcal{L}$, which can be stably trained by standard gradient optimization method like Adam \cite{kingma2014adam}.

% \section{\mname-SQL}
% \input{method}

\section{Applications}
% We apply DT-Fixup on two challenging NLP tasks: \emph{cross-domain} Text-to-SQL semantic parsing and multi-choice reading comprehension requiring \emph{logical reasoning}.
% Although inherently different, reasoning and structural understanding are crucial to perform well on both tasks.
%\vspace{-0.5em}

\subsection{Text-to-SQL Semantic Parsing}
% We now describe the task of cross-domain Text-to-SQL semantic parsing and our Text-to-SQL parser employed with DT-Fixup.
We first apply DT-Fixup on the task of cross-domain Text-to-SQL semantic parsing.
Given an \emph{unseen} schema $\mathcal{S}$ for a database during training, our goal is to translate the natural question $Q$ to the target SQL $T$.
The correct prediction depends on the interplay between the questions and the schema structures and the generalization over unseen schemas during inference.
As a result, reasoning and structural understanding are crucial to perform well on this task, especially for the more challenging cases.
We denote our baseline model as SQL-SP\footnote{\textbf{SQL}  \textbf{S}emantic \textbf{P}arser.} and henceforth.

%\vspace{-0.5em}

\paragraph{Implementation.}

% Given a schema $\mathcal{S}$ for a relational database, our goal is to translate the natural question $Q$ to the target SQL $T$.
% Here the question $Q=q_1\dots q_{\lvert Q \rvert}$ is a sequence of words, and the schema $\mathcal{S}=\{s_1,\dots, s_{\lvert \mathcal{S} \rvert}\}$ consists of tables and their columns. 
% $s \in \mathcal{S}$ can be either a table name or a column name containing words $s_{i,1},\dots,s_{i,\lvert s_i \rvert}$.
% Following \citet{wang2019rat}, a directed graph $\mathcal{G}=\langle\mathcal{V}, \mathcal{E}\rangle$ can be constructed to represent the relations between the inputs.
% Its nodes $\mathcal{V}=Q \cup \mathcal{S}$ include question tokens (each labeled with a corresponding token) and the columns and tables of the schema (each labeled with the words in its name).
% The edges $\mathcal{E}$ are defined following \citet{wang2019rat}.
% The target SQL $T$ is represented as an \emph{abstract syntax tree} in the context-free grammar of SQL.

For modeling Text-to-SQL generation, we adopt the \emph{encoder-decoder framework} which can be directly fit into the architecture shown in Fig.~\ref{fig:brto}. 
First, the pre-transformer module $f_e$ is a pre-trained language model which embeds the inputs $Q$ and $\mathcal{S}$ into joint representations $\pmb x_i$ for each column, table $s_i\in\mathcal{S}$ and question word $q_i\in Q$ respectively.
% Along with the relational embeddings $\pmb r^k, \pmb r^v$ specified by $\mathcal{G}$
The joint representations are passed into a sequence of $N$ relation-aware transformer layers.
The post-transformer module $f_o$ is a grammar-guided LSTM decoder, which uses the transformer output $\pmb y_i$ to predict the target SQL $T$.
% We use negative log-likelihood of the predicted SQL as the loss function $\mathcal{L}$.
We follow prior arts \cite{wang2019rat, guo2019towards, yin2018tranx} to implement SQL-SP.
The implementation details and hyperparameter settings are described in Appendix~\ref{sec:method}.

%\vspace{-0.5em}

\paragraph{Dataset.}

We evaluate SQL-SP on Spider \cite{yu2018spider}, a complex and cross-domain Text-to-SQL semantic parsing benchmark.
The dataset size is relatively small by deep learning standards, with only $10{,}181$ questions and $5{,}693$ queries covering $200$ databases in $138$ domains.

%\vspace{-0.5em}

\subsection{Logical Reading Comprehension}

The second task where we apply DT-Fixup is multi-choice reading comprehension requiring logical reasoning.
Given a context, a question and four options, the task is to select the right or most suitable answer.
Rather than extracting relevant information from a long context, this task relies heavily on the logical reasoning ability of the models.
% We denote our baseline model LMRC\footnote{\textbf{L}ogical \textbf{M}ulti-Choice \textbf{R}eading \textbf{C}omprehension.} as and henceforth.

%\vspace{-0.5em}

\paragraph{Implementation.}

% Similar to semantic parsing, our pre-transformer module $f_e$ is a pre-trained language model that embeds the input context, question and options. These representations are passed to a stack of $N$ vanilla transformer layers. Finally, the post-transformer module $f_o$ is a linear layer that predicts the correct option. 
On top of the pre-trained encodings of the input context, question and options, a stack of $N$ vanilla transformer layers are added before the final linear layer which gives the predictions.
The implementation details and hyperparamter settings are described in Appendix~\ref{sec:mrc}  

%\vspace{-0.5em}

\paragraph{Dataset.}

We evaluate on ReClor \cite{yu2020reclor}, a newly curated reading comprehension dataset requiring logical reasoning. The dataset contains logical reasoning questions taken from standardized exams (such as GMAT and LSAT) that are designed for students who apply for admission to graduate schools. Similar to Spider, this dataset is also small, with only $6{,}139$ questions.

\section{Experiments}
% \subsection{Experimental Setup}

% Our experiments are mainly conducted on the Spider~\cite{yu2018spider} benchmark.
% We use the database split proposed by \citet{yu2018spider} for evaluations, where $206$ databases are split into $146$ training, $20$ development and $40$ testing.
% % There are $8659$, $1034$, $2147$ question-SQL query pairs for training, development and testing.
% As the test set of Spider is only accessible through an evaluation server, most of our evaluations are performed on the development set.
% We use the exact match accuracy on all examples following \citet{yu2018spider}, which omits evaluation of generated values in the SQL queries.

% \mname-SQL is implemented in PyTorch. 
% We choose RoBERTa \cite{liu2019roberta} as the pre-trained language models.
% A sequence of 24 relation-aware transformer layers are stacked on top of $f_{\text{pre}}$.
% % For \mname, we choose the initial weight shrinkage parameter $\alpha$ as $15$.
% The Adam optimizer \cite{kingma2014adam} with the default hyperparameters is used to train the model with an initial learning rate $\eta$ of $4 \times 10^{-4}$.
% $\eta$ is annealed to $0$ with $4 \times 10^{-4}(1-steps/max\_steps)^{0.5}$.
% A separate learning rate is used to fine-tune the RoBERTa by multiplying $\eta$ a factor of $8 \times 10^{-3}$.
% We use a batch size of $16$ and train $60$ epochs (around $25,000$ steps).
% During inference, beam search is used with beam size as $5$.
% The complete hyperparameter configuration and how we tune it are presented in Appendix~\ref{sec:hyper}.

All the experiments in this paper are conducted with a signle 16GB Nvidia P100 GPU.
% The average training time is around 30 hours on Spider and 6 hours on ReClor.

\subsection{Semantic Parsing: Spider Results}

As the test set of Spider is only accessible through an evaluation server, most of our analyses are performed on the development set.
We use the exact match accuracy\footnote{We use the evaluation script provided in this repo: https://github.com/taoyds/spider} on all examples following \citet{yu2018spider}, which omits evaluation of generated values in the SQL queries.

\begin{table}[ht]
\centering
\small
\begin{tabular}{l l  l}
\hline
\bf Model & \bf Dev & \bf Test \\ \hline
% IRNet + BERT \cite{guo2019towards} & $61.9$ & $54.7$ \\
% RYANSQL + BERT \cite{choi2020ryansql} & $70.6$ & $60.6$ \\
\scriptsize RAT-SQL v3 + BERT \cite{wang2019rat} & $69.7$ & $65.6$\\
% \scriptsize Bridge + BERT (ensemble) \cite{lin2020bridging} & $71.1$ & $67.5$ \\
\scriptsize RAT-SQL + GraPPa \cite{yu2020grappa} & $73.4$ & $69.6$ \\
\scriptsize RAT-SQL + GAP \cite{shi2020learning} & $71.8$ & $69.7$  \\
\scriptsize RAT-SQL + GraPPa + GP \cite{zhao2021gp} & $72.8$ & $69.8$ \\
\scriptsize SGA-SQL + GAP (Anonymous) & $73.1$ & $70.1$ \\
\scriptsize RAT-SQL + GraPPa + Adv (Anonymous) & $75.5$ & $70.5$ \\
\hline
\scriptsize DT-Fixup SQL-SP + RobERTa (ours) & $75.0$ & $\mathbf{70.9}$ \\
\hline
\end{tabular}
\caption{Our accuracy on the Spider development and test sets, as compared to the other approaches at the top of the Spider leaderboard as of May $27$th, 2021.}
\label{tab:spider}
\end{table}

\newcommand{\xmark}{$\mathbin{\tikz [x=1.4ex,y=1.4ex,line width=.2ex] \draw (0,0) -- (1,1) (0,1) -- (1,0);}$}
\newcommand{\cmark}{$\checkmark$}
\begin{table}[ht]
\centering
\small
\begin{tabular}{l| c c  c |c}
\hline
\scriptsize \bf Model & \scriptsize $N$ &\scriptsize \bf Pretrain & \scriptsize \bf Epochs & \scriptsize \bf Acc. \\ \hline
\scriptsize RAT-SQL + BERT & $8$ & \xmark & $\sim 200$ & $69.7$\\
\scriptsize RAT-SQL + RoBERTa & $8$ & \xmark & $\sim 200$ & $69.6$ \\
\scriptsize RAT-SQL + GraPPa & $8$ & \cmark & $\sim 100$ & $73.4$ \\
\scriptsize RAT-SQL + GAP & $8$ & \cmark & $\sim 200$ & $71.8$ \\
\hline
\scriptsize SQL-SP + RoBERTa & $8$ & \xmark & $60$ & $66.9$ \\
\scriptsize + More Epochs & $8$ & \xmark & $100$ & $69.2$ \\
\scriptsize + DT-Fixup & $8$ & \xmark & $60$ & $73.5$ \\
\scriptsize + DT-Fixup \& More Layers & $24$ & \xmark & $60$ & $75.0$ \\ \hline
\scriptsize + T-Fixup$^*$ \& More Layers & $24$ & \xmark & $60$ & Failed \\
\hline
\end{tabular}
\caption{Comparisons with the models leveraging relational transformers on the Spider development set. {\bf Pretrain} here denotes task-specific pre-training, which leverges additional data and tasks, and is orthorgonal to our contribution. Not only we converge faster and reach better solution, simply training longer from the same baseline cannot close the performance gap. \change{$^*$We drop the constraints on the inputs to allow the application of T-Fixup in the mixed setup.}}
\label{tab:compare}
\end{table}

We present our results on the Spider leaderboard\footnote{https://yale-lily.github.io/spider} in Table~\ref{tab:spider}, where SQL-SP trained with DT-Fixup outperforms all the other approaches and achieves the new state of the art performance.
Notably, the top four submissions on the previous leaderboard are all occupied by models leveraging relation-aware transformers and task-specific pre-training.
Table~\ref{tab:compare} compares our proposed models with the publicly available works.
With enough training steps, our baseline model trained with the standard optimization strategy achieves the same level of performance as compared to RAT-SQL.
However, models trained with standard optimization strategy obtain much lower performance with the same epochs\footnote{One epoch iterates over the whole training set once. \citet{wang2019rat} trained with a batch size of $20$ for $90{,}000$ steps, which is around $200$ epochs on the Spider training set. \citet{yu2020grappa} trained with a batch size of $24$ for $40,000$ steps, which is around $100$ epochs on the Spider training set.} of training as compared to models trained with DT-Fixup and require more training steps to achieve the best accuracy.
At the same time, by adding more relation-aware transformer layers, further gains can be obtained for models trained with DT-Fixup, which achieves the state-of-the-art performance without any task-specific pre-training on additional data sources.
\change{As mentioned in Section~\ref{sec:tfixup}, in the mixed setup, there is no way to apply T-Fixup as it was originally proposed. The closest thing to compare is to drop its constraints on the inputs, but training then becomes highly unstable and fails to converge 4 times out of 5 runs.}
These results demonstrate the necessity and effectiveness of DT-Fixup to improve and accelerate the transformer training for Text-to-SQL parsers. 

\begin{table}[ht]
\centering
\scriptsize
\begin{tabular}{l c c c c c}
\hline
\bf Model & \bf Easy & \bf Medium & \bf Hard & \bf Extra & \bf All \\ \hline
\it Dev \\ 
RAT-SQL & $86.4$ &  $73.6$ & $62.1$ & $42.9$ & $69.7$\\
Bridge (ensemble) & $89.1$ & $71.7$ & $62.1$ & $\mathbf{51.8}$ & $71.1$\\ \hdashline
DT-Fixup SQL-SP & $91.9$ & $\mathbf{80.9}$ & $60.3$ & $48.8$ & $75.0$ \\ \hline
\it Test \\
RAT-SQL & $83.0$ &  $71.3$ & $58.3$ & $38.4$ & $65.6$\\
Bridge (ensemble) & $85.3$ & $73.4$ & $59.6$ & $40.3$ & $67.5$\\ \hdashline
DT-Fixup SQL-SP & $87.2$ & $\mathbf{77.5}$ & $60.9$ & $\mathbf{46.8}$ & $70.9$ \\
\hline
\end{tabular}
\caption{Breakdown of Spider accuracy by hardness.}
\label{tab:breakdown}
\end{table}

Table~\ref{tab:breakdown} shows the accuracy of our best model as compared to other approaches\footnote{We choose the top two submissions which also report the breakdown of the accuracy on the test set.} with different level of hardness defined by \citet{yu2018spider}.
We can see that a large portion of the improvement of our model comes from the medium level on both dev and test set.
Interestingly, while our model obtains similar performance for the extra hard level on the dev set, our model performs significantly better on the unseen test set.
As most of the extra hard cases involves implicit reasoning steps and complicated structures, it shows that our proposed models possess stronger reasoning and structural understanding ability, yielding better generalization over unseen domains and database schemas.

\subsection{Reading Comprehension: ReClor Results}

\begin{table}[ht]
\centering
\small
\begin{tabular}{l| c c}
\hline
\bf Model & \bf Dev & \bf Test \\ \hline
no extra layers$^*$ \cite{yu2020reclor} & $62.6$ & $55.6$\\ \hdashline
no extra layers & $63.6$ & $56.2$ \\
$4$ extra layers & $66.2$ & $58.2$ \\
$4$ extra layers + DT-Fixup & $\mathbf{66.8}$ & $\mathbf{61.0}$ \\ \hline
% RoBERTa + RACE & $68.0^*$ \\
% \hline
\end{tabular}
\caption{Our accuracy on ReClor. Star$^*$ is the best baseline model result reported in \cite{yu2020reclor} without using the additional RACE dataset \cite{lai2017race}.}
\label{tab:reclor_result}
\end{table}

\begin{table*}[ht!]
\begin{minipage}{.35\linewidth}
\centering
\small
\begin{tabular}{c  c  c}
\hline
\textbf{$N$} & \textbf{Standard} & \textbf{DT-Fixup} \\ \hline
\it Spider \\
2 & $69.47 \pm 0.30$ & $70.73 \pm 0.18$ \\
4 & $70.04 \pm 0.33$ & $72.22 \pm 0.61$\\
8 & $66.86 \pm 0.16$ & $73.24 \pm 0.51$\\
16 & $20.44 \pm 1.11$ & $73.52 \pm 0.47$ \\
24 & $19.37 \pm 0.16$ & $73.79 \pm 0.49$ \\
32 & $19.57 \pm 0.43$ & $73.02 \pm 0.52$\\
\hline
\it ReClor \\
4 & $64.05 \pm 0.44$ & $64.31 \pm 0.68$ \\
8 & $56.96 \pm 6.12$ & $65.31 \pm 0.62$ \\
16 & $27.10 \pm 1.50$ & $65.68 \pm 1.12$ \\ \hline
\end{tabular}
\captionof{table}{Ablation on the number of transformer layers $N$. The means and standard deviations are reported based on $5$ runs with different random seeds.}
\label{tab:layer}
\end{minipage}\hfill
\begin{minipage}{.63\linewidth}
  \centering
  \includegraphics[width=\linewidth]{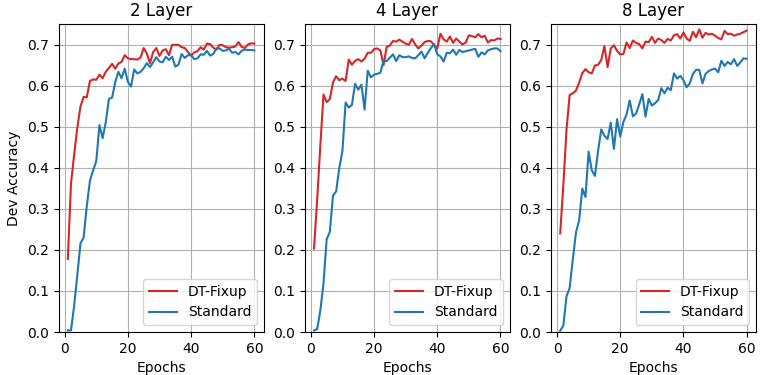}
  \captionof{figure}{Validation curves on Spider for models trained with different settings.}
  \label{fig:speed}
\end{minipage}
\end{table*}

\begin{table*}[h]
\begin{minipage}{.35\linewidth}
\footnotesize
\centering
%% \begin{tabular}{l | cccc |c }
%%   \hline
%%    Model & \bf Correct & \bf Column & \bf Sketch & \bf Both & \bf All \\ \hline
%% \bf Base & 39 & 51 & 92 & 124 & 306  \\
%% \bf Shallow & 35 & 60 & 83 & 105 & 283 \\
%% \bf Deep & 42 & 53 & 77 & 88 & 260 \\
%% \hline
%% \end{tabular}
\begin{tabular}{l | ccc}
  \hline
   & \bf Base & \bf Shallow & \bf Deep \\
  \hline  
\scriptsize{\bf False neg.\ } & 39 & 35 & 42 \\
\scriptsize{\bf Column err.\ only} & 51 & 60 & 53 \\
\scriptsize{\bf Sketch err.\ only} & 92 & 83 & 77 \\
\scriptsize{\bf Both err.\ } & 124 & 105 & 88 \\
\hline
\scriptsize{\bf All} & 306 & 283 & 260 \\
\hline
\end{tabular}
\captionof{table}{Failures in each category.}
\label{tab:error_all}
\end{minipage}%
\hfill
\begin{minipage}{.64\linewidth}
%\begin{figure}[h]
  \centering
  \begin{minipage}{.49\linewidth}
  \centering    
  \includegraphics[width=.95\linewidth]{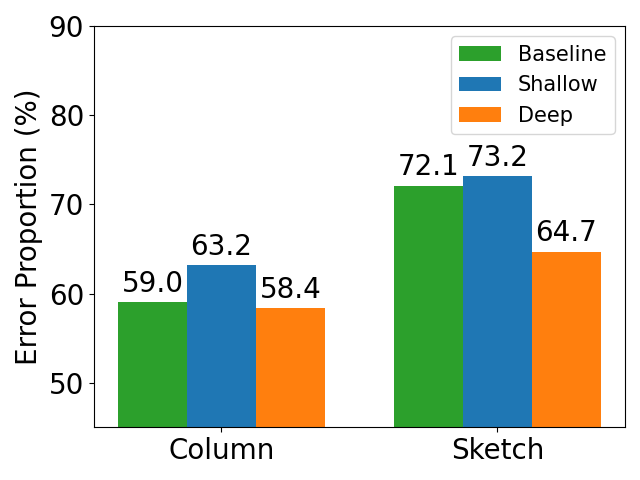}
  \captionof{figure}{Error breakdown on examples where \emph{all} models are wrong.}%
  \label{fig:err_all_m_wrong}
  \end{minipage}\hfill
  \begin{minipage}{.49\linewidth}
  \centering    
  \includegraphics[width=.95\linewidth]{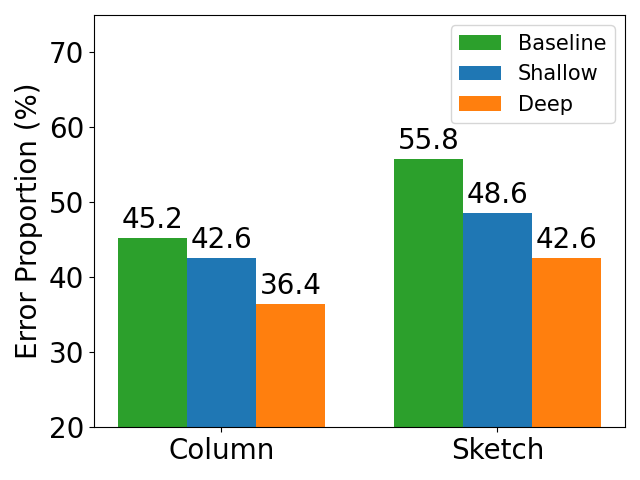}
  \captionof{figure}{Error breakdown on examples where \emph{any} model is wrong.}
  \label{fig:err_any_m_wrong}
  \end{minipage}
%\end{figure}
\end{minipage}
\end{table*}

% Similar to Spider, we evaluate the models with accuracy on the ReClor validation set, since the test set is not publicly available.
For ReClor, we choose the best model in \citet{yu2020reclor} as the baseline which employs a linear classifier on top of RoBERTa.
% By stacking another 4 vanilla transformer layers between RoBERTa and the linear classifier.
From the results presented in Table~\ref{tab:reclor_result}, we can see that simply stacking additional vanilla transformer layers outperforms the baseline and adding DT-Fixup further improves the accuracy, which ranks the second on the public leaderboard at the time of this submission\footnote{https://eval.ai/web/challenges/challenge-page/503/}.
The result further validates the benefit of adding extra transformer layers and the effectiveness of DT-Fixup.
% The second model is also provided in \citet{yu2020reclor} additionally uses pretraining on the RACE \cite{lai2017race} dataset. 

% Notably, LMRC trained with DT-Fixup also performs competitively with the second baseline without leveraging any task specific pretraining.

% Table~\ref{tab:reclor_layer} further validates the advantages of DT-Fixup over the standard optimization strategy when training deeper models. For all the experiments presented we use the same architecture and implementation and train the model for $12$ epochs. Training with the standard optimization strategy becomes unstable for models with 8 layers and completely fails for 16 layers. DT-Fixup, at the same time is successfully able to train deep models.

\subsection{Ablation Studies}
For fair comparisons and better understanding, we conduct multiple sets of ablation with the same architecture and implementation to validate the advantages of DT-Fixup over the standard optimization strategy.
Note that, the batch sizes in our experiments are relatively small (16 for Spider and 24 for ReClor) due to the size of the pre-trained models, while batch sizes for masked language modelling \cite{liu2019roberta} and machine translation \cite{huang2020improving} are commonly larger than $1024$.
% For all the experiments conducted on Spider in this section, we train the model for $60$ epochs.

% Yanshuai/Peng. This figures suggest there might be no difference if you just trained for longer.  I'd want to see more epochs.

\paragraph{Deeper Models.}

As we can see from Table~\ref{tab:layer}, the standard optimization strategy fails completely to train deep transformers whose depths are larger than $8$ on both Spider and ReClor, showing that it struggles to properly train the transformer model as the depth increases.
At the same time, DT-Fixup can successfully train deeper transformers up to $32$ layers and consistently achieves better performance than models trained by the standard optimization strategy with the same depth on both Spider and ReClor.
With DT-Fixup, deep models generally achieve better performance than the shallow ones even there are only thousands of training examples.
It contradicts the common belief that increasing depth of the transformer model is helpful only when there are enough training data.

\paragraph{Faster Convergence.}

Demonstrated by the validation curves on Spider plotted in Figure~\ref{fig:speed}, models trained with DT-Fixup converges to the same level of performance much faster than models trained with the standard optimization strategy.
While standard optimization strategy struggles as the models become deeper, DT-Fixup can keep the model training smooth, showing that DT-Fixup can effectively accelerate the convergence of the transformer training, especially for the deep ones.

\paragraph{Batch Sizes When Dataset Size is Small.}
\change{
As shown in Table~\ref{tab:bsz}, increasing batch size on Spider from 16 to 120, the average performance from five runs drops from 73.24 to 71.08 and the gap with the standard training approach becomes much narrower.
It empirically verifies that large-batch training has a negative impact on the generalization when the dataset size is small, confirming the need to stablize small batch training. 
}

\begin{table}[ht]
\centering
\small
\begin{tabular}{l| l c}
\hline
\bf Model & \bf Batch Size & \bf Acc \\ \hline
$8$ extra layers + Standard & 16 & $69.60\pm0.40$ \\
$8$ extra layers + DT-Fixup & 16 & $73.24\pm0.51$  \\ \hline
$8$ extra layers + DT-Fixup & 120 & $71.08\pm0.37$ \\
\hline
\end{tabular}
\caption{Ablation on the batch sizes for the Spider dataset. To enable large-batch training, we implement the trick of gradient accumulation at the expense of training speed. The means and standard deviations are reported based on 5 runs with different random seeds.}
\label{tab:bsz}
\end{table}

% \begin{table*}[ht]
% \centering
% \begin{tabular}{l | c c c c  | c}
% \hline
%  & \bf Easy & \bf Medium & \bf Hard & \bf Extra & \bf All \\ \hline
% \bf Base & 90.7 &  73.3 & 58.6 & 44.6 & 70.4\\
% \bf Shallow & 91.1 (+0.4) & 75.6 (+2.3) & 61.5 (+2.9) & 48.8 (+4.2) & 72.6 (+2.2)\\
% \bf Deep & 91.9 (+1.2) & 80.9 ({\bf + 7.6}) & 60.3 (+1.7) & 48.2 (+3.6) & 74.9 (+4.5) \\
% \hline
% \end{tabular}
% \caption{Breakdown of the accuracy by hardness for the three variants of our proposed model, where most of the improvement by increasing the depth comes from the medium level. These numbers are computed by the Spider evaluation script \cite{yu2018spider}.}
% \label{tab:breakdown}
% \end{table*}

\subsection{Source of the Improvements}

From the results on the Spider benchmark, we can see significant improvements by applying DT-Fixup and increasing the depth of the transformer model.
However, why and where they help Text-to-SQL semantic parsing are still unclear.
As an attempt to answer these questions, we investigate into the predicted results from three variants of our proposed model:
% Yanshuai/Peng Second sentence in the above paragraph needs some work. 
{\bf Baseline}, the best model ($N=4$) trained with the standard training approach;
{\bf Shallow}, a shallow model ($N=4$) trained with DT-Fixup;
{\bf Deep}, our best model ($N=24$) trained with DT-Fixup, which is much deeper.

% Table~\ref{tab:breakdown} shows the accuracy of these three models on examples with different level of hardness defined by \citet{yu2018spider}.
% We can see that impressive improvement from both applying DT-Fixup and increasing the number of layers.
% Interestingly, a significant portion of improvement by increasing the number of layers comes from the medium level of hardness, while applying DT-Fixup improves uniformly across all the level of hardness.

% However, the level of hardness defined by \citet{yu2018spider} may not be accurate in practice and the evaluation metric used for the Spider benchmark often leads to many false negatives as observed by \citet{wang2019rat}.
To better understand the models' behavior, we manually examine all the failed cases predicted by these models and classify the errors into four categories: 1) {\bf Correct:} equivalent in meaning but with different SQL syntax (\emph{e.g.}, \texttt{ORDER BY X LIMIT 1} and \texttt{SELECT MIN(X)}); 2) {\bf Column:}  the SQL structure is correct but there existed mispredicted columns; 3) {\bf Sketch:} the SQL structure is predicted different from the ground truth, while the aligned column prediction are correct;  4) {\bf Both:} there exist both sketch and column errors in the prediction.
Table~\ref{tab:error_all} presents the overall statistics of our error analysis.
Due to logically equivalent queries, there are a number of false negatives for all three models, confirming that the current Spider evaluation metric is not ideal.
At first glance, the improvements by applying DT-Fixup and increasing the depth seem to come from correcting {\bf Sketch} and {\bf Both} errors, while the three models make similar number of {\bf Column} only errors.
It provides evidence that applying DT-Fixup and increasing the depth can help the transformer model handle hard examples which are mispredicted completely (errors in {\bf Both} category) by the baseline model.
Typically, correct predictions on these hard examples require a certain level of reasoning and structural understanding ability.

\noindent \textbf{Fine-grained Error Analysis.}
In order to better understand the errors made, we look into the composition of error types by each model on mistaken examples common to all models, as well as on examples where at least one model is wrong. In Fig.\ \ref{fig:err_all_m_wrong}-\ref{fig:err_any_m_wrong}, ``Column'' means ``proportion with column errors'' (\emph{i.e.}, {\bf Column} or {\bf Both}); ``Sketch'' means ``proportion with sketch errors'' (\emph{i.e.}, {\bf Sketch} or {\bf Both}).
There are $190$ examples mispredicted by all the three models and $387$ examples which at least one of the three models mispredict.
% As shown in Figure~\ref{fig:acc}, the performance is roughly the same between the baseline and the shallow model.
Fig.\ \ref{fig:err_all_m_wrong}-\ref{fig:err_any_m_wrong} exclude false negatives due to equivalent logic queries, we can see the real improvements from the deep model are even more significant than what the exact match accuracy shows. Furthermore, among the common mistakes to all three models, the deep model has a much smaller proportion in the sketch mistakes which usually involve more logic and structure understanding. Some of column mistakes are due to missing domain knowledge or common sense, which is harder to improve without external data or knowledge. This shows that even among the failed cases, deeper transformer model can make more reasonable predictions.

\section{Related Work}

% \subsection*{Understanding Transformer Optimization}
Many research efforts have been devoted to understanding the training and improving the optimization of the transformer models.
In particular, transformer models often fail to learn unless a gradual learning rate warm-up is applied at the beginning of training.
\citet{chen2018best, nguyen2019transformers, wang2019learning} noticed a performance gap due to layer normalization, and introduced various architecture changes as remedy. \citet{zhang2019fixup, zhang2019improving, liu2020understanding} proposed initialization schemes to stabilize training, allowing either to remove layer normalization or learning rate warmup.
\citet{liu2019variance} demonstrated the instability of the Adam optimizer during early stages of optimization.
Based on these results, \citet{huang2020improving} proposed a weight initialization schema for the transformer that eliminates the need for layer normalization and warmup completely.

% \subsection*{Cross-Domain Text-to-SQL Semantic Parsing}
% Due to its potential to break the constraints for laymen to effectively access large databases through natural language, the task of translating natural questions to SQL queries attracts increasing research interest \cite{wang2019rat,guo2019towards,yu2020grappa,bogin2019global}.
% \citet{wang2019rat} introduce relational transformers to encode relational structure in the database schema and a given natural question, which improves the cross-domain schema generalization.
% Concurrently, various directions have been explored to improve cross-domain Text-to-SQL semantic parsing, including pre-training on synthetic data \cite{yu2020grappa}, improving the meaning representation \cite{guo2019towards}, leveraging graph neural networks \cite{bogin2019global} and so forth.

% \vspace{-0.5em}
\section{Conclusion}

Despite the broad applications of the transformer model, it struggles to perform well for some NLP tasks with limited training data.
In this work, we propose a theoretically justified optimization strategy DT-Fixup to train deeper transformer model with improved generalization and faster convergence speed on small datasets, which is generally applicable to different neural architectures.
On two important tasks, Text-to-SQL semantic parsing and logical reading comprehension that require reasoning and structural understanding, applying DT-Fixup achieves SOTA or near-SOTA results by simplying using extra transformer layers on top of the pre-trained models. Such observations suggest even boarder applicability of deeper transformers.

\section*{Acknowledgements}

We thank all the anonymous reviewers and area chair for their valuable inputs.

% Further analyses show that increasing the depth of the transformer model trained with limited data can be helpful for the generalization on complicated structural prediction tasks, instead of harmful as previously assumed.
% Such observations indicate that the current understanding of the transformer architecture is still incomplete and shed light on potential directions of future research.
\bibliographystyle{acl_natbib}
\bibliography{acl2021}

\begin{thebibliography}{32}
\expandafter\ifx\csname natexlab\endcsname\relax\def\natexlab#1{#1}\fi

\bibitem[{Ba et~al.(2016)Ba, Kiros, and Hinton}]{ba2016layer}
Jimmy~Lei Ba, Jamie~Ryan Kiros, and Geoffrey~E Hinton. 2016.
\newblock Layer normalization.
\newblock \emph{arXiv preprint arXiv:1607.06450}.

\bibitem[{Chen et~al.(2018)Chen, Firat, Bapna, Johnson, Macherey, Foster,
  Jones, Schuster, Shazeer, Parmar et~al.}]{chen2018best}
Mia~Xu Chen, Orhan Firat, Ankur Bapna, Melvin Johnson, Wolfgang Macherey,
  George Foster, Llion Jones, Mike Schuster, Noam Shazeer, Niki Parmar, et~al.
  2018.
\newblock The best of both worlds: Combining recent advances in neural machine
  translation.
\newblock In \emph{Proceedings of the 56th Annual Meeting of the Association
  for Computational Linguistics (Volume 1: Long Papers)}, pages 76--86.

\bibitem[{Devlin et~al.(2018)Devlin, Chang, Lee, and
  Toutanova}]{devlin2018bert}
Jacob Devlin, Ming-Wei Chang, Kenton Lee, and Kristina Toutanova. 2018.
\newblock Bert: Pre-training of deep bidirectional transformers for language
  understanding.
\newblock \emph{arXiv preprint arXiv:1810.04805}.

\bibitem[{Glorot and Bengio(2010)}]{glorot2010understanding}
Xavier Glorot and Yoshua Bengio. 2010.
\newblock Understanding the difficulty of training deep feedforward neural
  networks.
\newblock In \emph{Proceedings of the thirteenth international conference on
  artificial intelligence and statistics}, pages 249--256.

\bibitem[{Guo et~al.(2019)Guo, Zhan, Gao, Xiao, Lou, Liu, and
  Zhang}]{guo2019towards}
Jiaqi Guo, Zecheng Zhan, Yan Gao, Yan Xiao, Jian-Guang Lou, Ting Liu, and
  Dongmei Zhang. 2019.
\newblock Towards complex text-to-sql in cross-domain database with
  intermediate representation.
\newblock \emph{ACL}.

\bibitem[{Hochreiter and Schmidhuber(1997)}]{hochreiter1997long}
Sepp Hochreiter and J{\"u}rgen Schmidhuber. 1997.
\newblock Long short-term memory.
\newblock \emph{Neural computation}, 9(8):1735--1780.

\bibitem[{Huang et~al.(2020)Huang, P{\'e}rez, Ba, and
  Volkovs}]{huang2020improving}
Xiao~Shi Huang, Felipe P{\'e}rez, Jimmy Ba, and Maksims Volkovs. 2020.
\newblock Improving transformer optimization through better initialization.
\newblock \emph{ICML}.

\bibitem[{Keskar et~al.(2016)Keskar, Mudigere, Nocedal, Smelyanskiy, and
  Tang}]{keskar2016large}
Nitish~Shirish Keskar, Dheevatsa Mudigere, Jorge Nocedal, Mikhail Smelyanskiy,
  and Ping Tak~Peter Tang. 2016.
\newblock On large-batch training for deep learning: Generalization gap and
  sharp minima.
\newblock \emph{arXiv preprint arXiv:1609.04836}.

\bibitem[{Kingma and Ba(2014)}]{kingma2014adam}
Diederik~P Kingma and Jimmy Ba. 2014.
\newblock Adam: A method for stochastic optimization.
\newblock \emph{arXiv preprint arXiv:1412.6980}.

\bibitem[{Lai et~al.(2017)Lai, Xie, Liu, Yang, and Hovy}]{lai2017race}
Guokun Lai, Qizhe Xie, Hanxiao Liu, Yiming Yang, and Eduard Hovy. 2017.
\newblock {RACE}: Large-scale {R}e{A}ding comprehension dataset from
  examinations.
\newblock In \emph{Proceedings of the 2017 Conference on Empirical Methods in
  Natural Language Processing}, pages 785--794, Copenhagen, Denmark.
  Association for Computational Linguistics.

\bibitem[{Lan et~al.(2019)Lan, Chen, Goodman, Gimpel, Sharma, and
  Soricut}]{lan2019albert}
Zhenzhong Lan, Mingda Chen, Sebastian Goodman, Kevin Gimpel, Piyush Sharma, and
  Radu Soricut. 2019.
\newblock Albert: A lite bert for self-supervised learning of language
  representations.
\newblock \emph{arXiv preprint arXiv:1909.11942}.

\bibitem[{Liu et~al.(2019{\natexlab{a}})Liu, Jiang, He, Chen, Liu, Gao, and
  Han}]{liu2019variance}
Liyuan Liu, Haoming Jiang, Pengcheng He, Weizhu Chen, Xiaodong Liu, Jianfeng
  Gao, and Jiawei Han. 2019{\natexlab{a}}.
\newblock On the variance of the adaptive learning rate and beyond.
\newblock \emph{arXiv preprint arXiv:1908.03265}.

\bibitem[{Liu et~al.(2020)Liu, Liu, Gao, Chen, and Han}]{liu2020understanding}
Liyuan Liu, Xiaodong Liu, Jianfeng Gao, Weizhu Chen, and Jiawei Han. 2020.
\newblock Understanding the difficulty of training transformers.
\newblock \emph{EMNLP}.

\bibitem[{Liu et~al.(2019{\natexlab{b}})Liu, Ott, Goyal, Du, Joshi, Chen, Levy,
  Lewis, Zettlemoyer, and Stoyanov}]{liu2019roberta}
Yinhan Liu, Myle Ott, Naman Goyal, Jingfei Du, Mandar Joshi, Danqi Chen, Omer
  Levy, Mike Lewis, Luke Zettlemoyer, and Veselin Stoyanov. 2019{\natexlab{b}}.
\newblock Roberta: A robustly optimized bert pretraining approach.
\newblock \emph{arXiv preprint arXiv:1907.11692}.

\bibitem[{Nguyen and Salazar(2019)}]{nguyen2019transformers}
Toan~Q Nguyen and Julian Salazar. 2019.
\newblock Transformers without tears: Improving the normalization of
  self-attention.
\newblock \emph{arXiv preprint arXiv:1910.05895}.

\bibitem[{Popel and Bojar(2018)}]{popel2018training}
Martin Popel and Ond{\v{r}}ej Bojar. 2018.
\newblock Training tips for the transformer model.
\newblock \emph{The Prague Bulletin of Mathematical Linguistics},
  110(1):43--70.

\bibitem[{Radford et~al.(2019)Radford, Wu, Child, Luan, Amodei, and
  Sutskever}]{radford2019language}
Alec Radford, Jeffrey Wu, Rewon Child, David Luan, Dario Amodei, and Ilya
  Sutskever. 2019.
\newblock Language models are unsupervised multitask learners.
\newblock \emph{OpenAI Blog}, 1(8):9.

\bibitem[{Shaw et~al.(2018)Shaw, Uszkoreit, and Vaswani}]{shaw2018self}
Peter Shaw, Jakob Uszkoreit, and Ashish Vaswani. 2018.
\newblock Self-attention with relative position representations.
\newblock In \emph{Proceedings of the 2018 Conference of the North American
  Chapter of the Association for Computational Linguistics: Human Language
  Technologies, Volume 2 (Short Papers)}, pages 464--468.

\bibitem[{Shi et~al.(2020)Shi, Ng, Wang, Zhu, Li, Wang, Santos, and
  Xiang}]{shi2020learning}
Peng Shi, Patrick Ng, Zhiguo Wang, Henghui Zhu, Alexander~Hanbo Li, Jun Wang,
  Cicero Nogueira~dos Santos, and Bing Xiang. 2020.
\newblock Learning contextual representations for semantic parsing with
  generation-augmented pre-training.
\newblock \emph{arXiv preprint arXiv:2012.10309}.

\bibitem[{Srivastava et~al.(2014)Srivastava, Hinton, Krizhevsky, Sutskever, and
  Salakhutdinov}]{srivastava2014dropout}
Nitish Srivastava, Geoffrey Hinton, Alex Krizhevsky, Ilya Sutskever, and Ruslan
  Salakhutdinov. 2014.
\newblock Dropout: a simple way to prevent neural networks from overfitting.
\newblock \emph{The journal of machine learning research}, 15(1):1929--1958.

\bibitem[{Szegedy et~al.(2016)Szegedy, Vanhoucke, Ioffe, Shlens, and
  Wojna}]{szegedy2016rethinking}
Christian Szegedy, Vincent Vanhoucke, Sergey Ioffe, Jon Shlens, and Zbigniew
  Wojna. 2016.
\newblock Rethinking the inception architecture for computer vision.
\newblock In \emph{Proceedings of the IEEE conference on computer vision and
  pattern recognition}, pages 2818--2826.

\bibitem[{Vaswani et~al.(2017)Vaswani, Shazeer, Parmar, Uszkoreit, Jones,
  Gomez, Kaiser, and Polosukhin}]{vaswani2017attention}
Ashish Vaswani, Noam Shazeer, Niki Parmar, Jakob Uszkoreit, Llion Jones,
  Aidan~N Gomez, {\L}ukasz Kaiser, and Illia Polosukhin. 2017.
\newblock Attention is all you need.
\newblock In \emph{Advances in neural information processing systems}, pages
  5998--6008.

\bibitem[{Wang et~al.(2019{\natexlab{a}})Wang, Shin, Liu, Polozov, and
  Richardson}]{wang2019rat}
Bailin Wang, Richard Shin, Xiaodong Liu, Oleksandr Polozov, and Matthew
  Richardson. 2019{\natexlab{a}}.
\newblock Rat-sql: Relation-aware schema encoding and linking for text-to-sql
  parsers.
\newblock \emph{arXiv preprint arXiv:1911.04942}.

\bibitem[{Wang et~al.(2019{\natexlab{b}})Wang, Li, Xiao, Zhu, Li, Wong, and
  Chao}]{wang2019learning}
Qiang Wang, Bei Li, Tong Xiao, Jingbo Zhu, Changliang Li, Derek~F Wong, and
  Lidia~S Chao. 2019{\natexlab{b}}.
\newblock Learning deep transformer models for machine translation.
\newblock In \emph{Proceedings of the 57th Annual Meeting of the Association
  for Computational Linguistics}, pages 1810--1822.

\bibitem[{Xu et~al.(2019)Xu, Liu, van Genabith, Xiong, and
  Zhang}]{xu2019lipschitz}
Hongfei Xu, Qiuhui Liu, Josef van Genabith, Deyi Xiong, and Jingyi Zhang. 2019.
\newblock Lipschitz constrained parameter initialization for deep transformers.
\newblock \emph{arXiv preprint arXiv:1911.03179}.

\bibitem[{Yin and Neubig(2018)}]{yin2018tranx}
Pengcheng Yin and Graham Neubig. 2018.
\newblock Tranx: A transition-based neural abstract syntax parser for semantic
  parsing and code generation.
\newblock \emph{arXiv preprint arXiv:1810.02720}.

\bibitem[{Yu et~al.(2020{\natexlab{a}})Yu, Wu, Lin, Wang, Tan, Yang, Radev,
  Socher, and Xiong}]{yu2020grappa}
Tao Yu, Chien-Sheng Wu, Xi~Victoria Lin, Bailin Wang, Yi~Chern Tan, Xinyi Yang,
  Dragomir Radev, Richard Socher, and Caiming Xiong. 2020{\natexlab{a}}.
\newblock Grappa: Grammar-augmented pre-training for table semantic parsing.
\newblock \emph{arXiv preprint arXiv:2009.13845}.

\bibitem[{Yu et~al.(2018)Yu, Zhang, Yang, Yasunaga, Wang, Li, Ma, Li, Yao,
  Roman et~al.}]{yu2018spider}
Tao Yu, Rui Zhang, Kai Yang, Michihiro Yasunaga, Dongxu Wang, Zifan Li, James
  Ma, Irene Li, Qingning Yao, Shanelle Roman, et~al. 2018.
\newblock Spider: A large-scale human-labeled dataset for complex and
  cross-domain semantic parsing and text-to-sql task.
\newblock In \emph{Proceedings of the 2018 Conference on Empirical Methods in
  Natural Language Processing}, pages 3911--3921.

\bibitem[{Yu et~al.(2020{\natexlab{b}})Yu, Jiang, Dong, and
  Feng}]{yu2020reclor}
Weihao Yu, Zihang Jiang, Yanfei Dong, and Jiashi Feng. 2020{\natexlab{b}}.
\newblock Reclor: A reading comprehension dataset requiring logical reasoning.
\newblock \emph{arXiv preprint arXiv:2002.04326}.

\bibitem[{Zhang et~al.(2019{\natexlab{a}})Zhang, Titov, and
  Sennrich}]{zhang2019improving}
Biao Zhang, Ivan Titov, and Rico Sennrich. 2019{\natexlab{a}}.
\newblock Improving deep transformer with depth-scaled initialization and
  merged attention.
\newblock In \emph{Proceedings of the 2019 Conference on Empirical Methods in
  Natural Language Processing and the 9th International Joint Conference on
  Natural Language Processing (EMNLP-IJCNLP)}, pages 897--908.

\bibitem[{Zhang et~al.(2019{\natexlab{b}})Zhang, Dauphin, and
  Ma}]{zhang2019fixup}
Hongyi Zhang, Yann~N Dauphin, and Tengyu Ma. 2019{\natexlab{b}}.
\newblock Fixup initialization: Residual learning without normalization.
\newblock \emph{ICLR}.

\bibitem[{Zhao et~al.(2021)Zhao, Cao, and Zhao}]{zhao2021gp}
Liang Zhao, Hexin Cao, and Yunsong Zhao. 2021.
\newblock Gp: Context-free grammar pre-training for text-to-sql parsers.
\newblock \emph{arXiv preprint arXiv:2101.09901}.

\end{thebibliography}

%\appendix

\newpage
\appendix

\onecolumn
\vspace{-.5em}
\section{Full Proof}
\vspace{-.5em}
\label{sec:full_proof}
\begin{customthm}{3.1}
  Assuming $\norm{\px} = \Theta(\mu)$ for some $\mu \gg 1$, then $\norm{\pd{f_G}{\ptheta_G}} = \Theta(1)$ if $\norm{v_l} = \norm{w_l} = \norm{\rv_l} = \Theta{\left(((4\mu^2 + 2\mu + 2)N)^{-\frac{1}{2}}\right)}$ for all encoder layers $l$ in relational transformers; and $\norm{v_l} = \norm{w_l} = \Theta{\left((4\mu^2N)^{-\frac{1}{2}}\right)}$ in the case of vanilla transformers.
\end{customthm}

{\it Proof.}
First, let's inspect the feedforward pass through the transformer blocks, which have nonlinear layers $G_l$'s and skip connections:
  $\px_1 = \px ; ~~~~ \px_2  = \px_1 + G_1(\px_1, \ptheta_1); ~~\ldots; ~~ \px_{l+1}  = \px_l + G_l(\px_l, \ptheta_l)$
For $l \% 2 = 1$ (i.e. odd layers), $G_l$ is a (relational) self-attention layer, whereas for even layers, $G_l$ is a MLP layer.
Using $\normeq$ to denote bounded in norm as in \citet{huang2020improving}, then at initialization:
\vspace{-.5em}
\begin{align}
  \px_{l+1}  &\normeq \px_l + v_lw_l\px_l + w_l\rv_l \quad\quad\quad\quad\text{For relational self-attention}\label{px_l_recur_rat} \\
  \px_{l+1}  &\normeq \px_l + v_lw_l\px_l \quad\quad\quad\quad\quad\quad\quad\text{For vanilla self-attention and MLP}\label{px_l_recur_mlp} 
\end{align}
This is due to the fact that the probability from softmax sums to one, so does not alter the overall norm; at initialization, values are at the linear identity range of the nonlinearities.
Therefore, for all three types of layers: $\pd{\px_{l+1}}{\px_l} \normeq 1 + v_lw_l$ and $\pd{G_l}{\px_l} \normeq v_lw_l$.
And for relational self-attention: $\pd{\px_{l+1}}{\ptheta_{l}} =  \pd{G_l}{\ptheta_{l}} \normeq [w_l\px_l, v_l\px_l + \rv_l, w_l, \pmb 0]$,
where $\pmb 0$ are due to $q$, $k$, $\prk$ which appear only inside the softmax and do not asymptotically affect the norm.
And for vanilla self-attention and MLP, $\pd{\px_{l+1}}{\ptheta_{l}} =  \pd{G_l}{\ptheta_{l}} \normeq [w_l\px_l, v_l\px_l, \pmb 0]$.
Next, let's look at $\pd{f_G}{\ptheta_G} = [\pd{f_G}{\ptheta_1},\ldots,\pd{f_G}{\ptheta_l},\ldots,\pd{f_G}{\ptheta_L}]$. First note that:
\vspace{-.5em}
\begin{equation}
f_G(\px, \ptheta_G) = \px_1 + G_1(\px_1, \ptheta_1) + G_2(\px_2, \ptheta_2) + \ldots + G_L(\px_2, \ptheta_L)
\end{equation}
Working backwards, for the last layer, $\pd{f_G}{\ptheta_L}\! =\! \pd{G_L}{\ptheta_L}$. For $\pd{f_G}{\ptheta_l}$, terms with index lower than $l$ vanish, so:
\vspace{-.5em}
\begin{align}
  \pdi{f_G}{\ptheta_l} &= \pdi{G_l}{\ptheta_l} + \pdi{G_{l+1}}{\px_{l+1}}\pdi{\px_{l+1}}{\ptheta_l} + \ldots + \pdi{G_L}{\px_L}\pdi{\px_{L}}{\px_{L-1}}\ldots\pdi{\px_{l+1}}{\ptheta_l} \\
  \phantom{\pdi{f_G}{\ptheta_l}} &\normeq \left(1 + v_{l+1}w_{l+1} + \ldots + v_{L}w_{L} (1+v_{L-1}w_{L-1})\ldots(1+v_{l+1}w_{l+1})\right) \pdi{G_l}{\ptheta_l}
\end{align} \vspace{-.5em}
Assuming $v_1 \normeq v_2 \ldots \normeq v_L$ and $w_1 \normeq w_2 \ldots  \normeq w_L$, and both $\ll 1$, then the above reduces to:
\vspace{-.5em}
\begin{equation}
  \pdi{f_G}{\ptheta_l}  \normeq (1 + (L - l) v_lw_l) \pdi{G_l}{\ptheta_l}
\end{equation}
Recall that we want to bound $\pd{f_G}{\ptheta_G} \pd{f_G}{\ptheta_G}^{\top}= {\textstyle \sum}_l \pd{f_G}{\ptheta_l}\pd{f_G}{\ptheta_l}^{\top}$. For vanilla self-attention or MLP layers:
\vspace{-.5em}
\begin{align}
\pd{f_G}{\ptheta_l}\pd{f_G}{\ptheta_l}^{\top} \normeq \left(\norm{w_l}^2\norm{\px_l}^2 + \norm{v_l}^2\norm{\px_l}^2\right) (1 + (L - l)\norm{v_l}\norm{w_l})^2 \label{f_theta_mlp}
\end{align}%
And for relational self-attention: %
\vspace{-.2em}
\begin{align}
\pd{f_G}{\ptheta_l}\pd{f_G}{\ptheta_l}^{\top} \!\!\normeq\!\! \left(\norm{w_l}^2\norm{\px_l}^2 \!\!+\!\! \norm{v_l}^2\norm{\px_l}^2 \!\!+\!\! 2\norm{v_l}\norm{\px_l}\norm{\rv_l} \!\!+\!\! \norm{\rv_l}^2 \!\!+\!\! \norm{w_l}^2\right) (1 \!\!+\!\! (L \!-\! l)\norm{v_l}\norm{w_l})^2 \label{f_theta_rat}
\end{align}
At initialization, we want $v_l$, $w_l$, $\rv_l$ of all layers to have the same norm, i.e. $\norm{v_l} \normeq \norm{w_l} \normeq \norm{\rv_l} \normeq \norm{v_j} \normeq \norm{w_j} \normeq \norm{\rv_j}$ for all $l$ and $j$, so denoting them using $\xi$. And recall that $N$ is the number of transformer blocks, with each block containing two layers, so that $2N = L$. So we have:%
\vspace{-.5em}
\begin{align}
  \pd{f_G}{\ptheta_G}\pd{f_G}{\ptheta_G}^{\top} &\!\!\normeq\! {\textstyle \sum}_{l\%2=0}\!\left(\!2\xi^2 \norm{\px_l}^2\!\right)\left(1\!+\!(L\!-\!l)\xi^2\right) \!+\! {\textstyle \sum}_{l\%2=1}\! \left(\! 2\xi^2 \norm{\px_l}^2 \!+\! 2\xi^2 \norm{\px_l} \!+\! 2\xi^2\!\right)\!\left(\!1\!+\!(L\!-\!l)\xi^2\right) \nonumber \\ 
  \phantom{\pd{f_G}{\ptheta_G}\pd{f_G}{\ptheta_G}^{\top}} &\normeq {\textstyle \sum}^N_{l=1} \left( 4\xi^2 \norm{\px_l}^2 + 2\xi^2\norm{\px_l} + 2\xi^2 \right)(1 + (2N - l)\xi^2 ) \label{f_g2_theta_g2_final}
\end{align}%
Similarly if $f_G$ is vanilla transformer instead of a relational one, we have:
\vspace{-.5em}
\begin{align}
  \pd{f_G}{\ptheta_G}\pd{f_G}{\ptheta_G}^{\top} &\normeq {\textstyle \sum}^N_{l=1} \left( 4\xi^2 \norm{\px_l}^2 \right)(1 + (2N - l)\xi^2 ) \label{f_g2_theta_g2_final_vanilla}
\end{align}
%%   \pd{f_G}{\ptheta_G}\pd{f_G}{\ptheta_G}^{\top} &\normeq {\textstyle \sum}_{l\%2=0} \left( 2\xi^2 (1+\xi^2)^{2l}\norm{\px}^2\right)\left(1+(L-l)\xi^2\right) \nonumber \\
%% \phantom{\pd{f_G}{\ptheta_G}\pd{f_G}{\ptheta_G}^{\top}}  &\phantom{\normeq} + {\textstyle \sum}_{l\%2=1} \left( 4\xi^2 (1+\xi^2)^{2l}\norm{\px}^2 + l\xi^2+ 2\xi^2\right)\left(1+(L-l)\xi^2\right)
The only variable that still depends on $l$ is $\px_l$, which by expanding the recursion in Eq.\ \ref{px_l_recur_rat}-\ref{px_l_recur_mlp}, gives:
\begin{align}
  \px_{l}  &\normeq (1+\xi^2)^l \px \normeq (1 + l\xi^2 + \Theta(\xi^4))\px \quad\quad\quad\quad\quad\quad\quad\quad\quad \text{For vanillla transformer}\label{pxl_vanilla} \\
  \px_{l}  &\normeq (1+\xi^2)^l \px + l/2\xi^2 \normeq (1 + l\xi^2 + \Theta(\xi^4))\px + l/2\xi^2 \quad \text{For relational transformer} \label{pxl_rat}
\end{align}
Now let $\norm{\px} \normeq \mu$ , and we have assumed that $\mu \gg 1$, which is very common for output of pre-trained encoders, and due to the high dimensionality. And let %
\vspace{-.3em}
\begin{equation}
  \xi = \left(N (4\mu^2 + 2\mu + 2)\right)^{-\frac{1}{2}} \label{xi_def}
\end{equation}%
Then substituting it into Eq.\ \ref{pxl_vanilla}-\ref{pxl_rat}, we have $\px_{l} \normeq \px$ for all types of layers.
Similarly, plugging Eq.\ \ref{xi_def} into the expression $(1 + (2N - l)\xi^2 )$ in Eq.\ \ref{f_g2_theta_g2_final} yields $(1 + (2N - l)\xi^2 ) \normeq 1$, together with $\px_{l} \normeq \px$, and Eq.\ \ref{xi_def}, Eq.\ \ref{f_g2_theta_g2_final} becomes: %
\vspace{-.3em}
\begin{equation*}
  \pd{f_G}{\ptheta_G}\pd{f_G}{\ptheta_G}^{\top} \normeq {\textstyle \sum}^N_{l=1} \frac{4 \mu^2}{N \left(4\mu^2 + 2 \mu + 2\right)} + \frac{2\mu}{N \left(4\mu^2 + 2 \mu + 2\right)} + \frac{2}{N \left(4\mu^2 + 2 \mu + 2\right)} \normeq {\textstyle \sum}^N_{l=1} 1/N = \Theta(1)
\end{equation*}%
\vspace{-.3em}
This concludes the proof for relational transformers. For vanilla transformers, with $\xi = \left(N (4\mu^2)\right)^{-\frac{1}{2}}$, and following the same steps, but plugging into Eq.\ \ref{f_g2_theta_g2_final_vanilla}, we have $\pd{f_G}{\ptheta_G}\pd{f_G}{\ptheta_G}^{\top} \normeq 1$. \hfill Q.E.D.

\vspace{-.5em}
\section{Proof of Theorem~\ref{theorem:main}}
\vspace{-.5em}
\label{sec:proof}
For brevity, we drop the layer index. But for the relation embeddings, for clarity, we will consider the individual components of $\prv,\prk$ instead of considering the scalar case. 

{\it Proof.} We will focus the self-attention layer, as the skip connection and MLP layers are analyzed in \citet{huang2020improving}. As mentioned in the main text, since what we care is the magnitude of the update, we assume $d_x=1$ and drop layer index $l$ without loss of generality.
In this case, the projection matrices $\pmb q, \pmb k, \pmb v, \pmb w$  reduce to scalars $q, k, v, w\in \mathbb{R}$. The input $\pmb x$ and the relational embeddings $\pmb r^k, \pmb r^v$ are $n\times1$ vectors. 
For a single query input $x' \in \pmb x$, the attention layer (without skip connection) is defined as follows:
\vspace{-.5em}
\begin{equation*}
    G(x')  = \text{softmax}\left(\frac1{\sqrt{d_x}}x' q(k \pmb x + \pmb r^k)^\top\right)(\pmb x v + \pmb r^v)w = {\textstyle \sum}_{i=1}^n \frac{e^{x'q(kx_i+r_i^k)}}{{\textstyle \sum}_{j=1}^ne^{x'q(kx_j+r_j^k)}} (x_iv+r_i^v)w 
\end{equation*}
Note that we are abusing the notation and take $G$ to be just the self-attention layer output here.
Let $s_i=e^{x'q(kx_i+r_i^k)}/{\textstyle \sum}_{j=1}^n e^{x'q(kx_j+r_j^k)}$ and $\delta_{ij}=1$ if $i=j$ and 0 otherwise, we can get: %
\begin{equation*}
    \begin{split}
        \pdi{G}{k} &= x'qw{\textstyle \sum}_{i=1}^n(x_iv+r_i^v)s_i\left(x_i-{\textstyle \sum}_{j=1}^nx_js_j\right) \\
        \pdi{G}{q} &= x'w{\textstyle \sum}_{i=1}^n(x_iv+r_i^v)s_i\left(kx_i + r_i^k -{\textstyle \sum}_{j=1}^n(kx_j+r_j^k)s_j\right)\\
        \pdi{G}{r_i^k} &= x'qw\left(-(x_i v + r_i^v)s_i + {\textstyle \sum}_{j=1}^n(x_jv+r_j^v)s_j\right)~; ~~~\pdi{G}{v} = w{\textstyle \sum}_{i=1}^nx_is_i \\         
        \pdi{G}{w} &= {\textstyle \sum}_{i=1}^n(x_iv+r_i^v)s_i ~; ~~~ \pdi{G}{r_i^v} = ws_i ~; ~~~ \pdi{G}{x_i} = vws_i + w{\textstyle \sum}_{j=1}^n\pdi{s_j}{x_i}(x_jv+r_j^v)
        % &= vws_i + w{\textstyle \sum}_{j=1}^n(x_jv+r_j^v)s_j(\delta_{ij}-s_i)x'qk
        % &= vws_i + wx'qks_i\left(x_iv+r_i^v - {\textstyle \sum}_{j=1}^n(x_jv+r_j^v)s_j\right)
    \end{split}
\end{equation*}%
\noindent When $x_i \ne x'$, we have: $\pd{s_j}{x_i} = s_j(\delta_{ij}-s_i)x'qk$; 
\noindent When $x_i = x'$, we have: $\pd{s_j}{x_i} = q\left((1 + \delta_{ij})kx_i+r_i^k\right)s_j - {\textstyle \sum}_{t=1}^nq\left((1+\delta_{it})kx_t+r_t^k\right)s_js_t$
Using Taylor expansion, we get that the SGD update $\Delta G$ is proportional to the magnitude of the gradient:%
\begin{equation*}
    \begin{split}
    \Delta G &= -\eta\pd{ \mathcal{L}}{ G}\left(\pd{ G}{ k}\pd{ G}{ k}^\top    
    + \pd{ G}{ q}\pd{ G}{ q}^\top 
    + \pd{ G}{ v}\pd{ G}{ v}^\top
    + \pd{ G}{ w}\pd{ G}{ w}^\top \right.\\
    & \left. + {\textstyle \sum}_{i=1}^n\pd{ G}{ r_i^k}\pd{ G}{ r_i^k}^\top
    + {\textstyle \sum}_{i=1}^n\pd{ G}{ r_i^v}\pd{ G}{ r_i^v}^\top 
    + {\textstyle \sum}_{i=1}^n\pd{ G}{ x_i}\pd{ G}{ x_i}^\top \right) + O(\eta^2)
    \end{split}
\end{equation*}%
\noindent By the assumption that $\lVert \eta \pd{ \mathcal{L}}{ G} \rVert = \Theta(\eta)$, we need to bound the term inside the main parentheses by $\Theta(1/L)$.
The desired magnitude $\Theta(1/L)$ is smaller than 1 so terms with lower power are dominating.
With $s_i\ge0$ and ${\textstyle \sum} s_i=1$, the following terms have the lowest power inside the main parentheses:%
\begin{equation*}
    \small
    \begin{split}
        \pd{ G}{ v}\pd{ G}{ v}^\top &= w^2({\textstyle \sum}_{i=1}^nx_is_i)^2=\Theta(\lVert w \rVert^2 \lVert x_i \rVert^2), \ i=1, \dots, n\\
        \pd{ G}{ w}\pd{ G}{ w}^\top &= ({\textstyle \sum}_{i=1}^n(x_iv+r_i^v)s_i)^2=\Theta(\lVert v \rVert^2 \lVert x_i \rVert^2) + 2\Theta(\lVert v \rVert \lVert r_i^v\rVert \lVert x_i \rVert ) + \Theta(\lVert r_i^v\rVert^2), \ i = 1, \dots, n\\
        {\textstyle \sum}_{i=1}^n\pd{ G}{ r_i^v}\pd{ G}{ r_i^v}^\top & = w^2{\textstyle \sum}_{i=1}^ns_i^2 =\Theta(\lVert w \rVert^2).\\
    \end{split}
\end{equation*}
\noindent For the MLP layer, all terms related to $r_i^v$ disappear, including the single $\Theta(\lVert w \rVert^2)$ in the last row.
By combining the update norm terms from both the self-attention and the MLP layers give the result. \hfill \textbf{Q.E.D.}
Note: The above theorem and analysis applies to a single layer, not the whole transformer module of many layers. In order to derive the scaling factor, one needs ensure that the output scale for each block is bounded by its input scale. This indeed holds for our scheme, but the complete proof is in Sec.\ \ref{sec:full_proof}.
\vspace{-.5em}
\section{Implementation Details of SQL-SP}
\label{sec:method}
\vspace{-.5em}
Given a schema $\mathcal{S}$ for a relational database, our goal is to translate the natural question $Q$ to the target SQL $T$.
Here the question $Q=q_1\dots q_{\lvert Q \rvert}$ is a sequence of words, and the schema $\mathcal{S}=\{s_1,\dots, s_{\lvert \mathcal{S} \rvert}\}$ consists of tables and their columns. 
$s \in \mathcal{S}$ can be either a table name or a column name containing words $s_{i,1},\dots,s_{i,\lvert s_i \rvert}$.
Following \citet{wang2019rat}, a directed graph $\mathcal{G}=\langle\mathcal{V}, \mathcal{E}\rangle$ can be constructed to represent the relations between the inputs.
Its nodes $\mathcal{V}=Q \cup \mathcal{S}$ include question tokens (each labeled with a corresponding token) and the columns and tables of the schema (each labeled with the words in its name).
The edges $\mathcal{E}$ are defined following \citet{wang2019rat}.
The target SQL $T$ is represented as an \emph{abstract syntax tree} in the context-free grammar of SQL.
\vspace{-.5em}
\subsection{Encoder}
\vspace{-.5em}
Following \cite{wang2019rat, guo2019towards}, our pre-transformer module $f_e$ leverages pre-trained language models to obtain the input $X$ to the main transformer module. 
First, the sequence of words in the question $Q$ are concatenated with all the items (either a column or a table) in the schema $\mathcal{S}$.
In order to prevent our model from leveraging potential spurious correlations based on the order of the items, the items in the schema are concatenated in random order during training.
We feed the concatenation into the pre-trained model and extract the last hidden states $\pmb x_i^{(q)}$ and $\pmb h_i = \pmb h_{i, 1}, \dots, \pmb h_{i, \lvert s_i \rvert}$ for each word in $Q$ and each item in $\mathcal{S}$ respectively.
For each item $s_i$ in the schema, we run an additional bidirectional LSTM (BiLSTM) \cite{hochreiter1997long} over the hidden states of the words in its name $\pmb h_i$.
We then add the average hidden state and the final hidden state of the BiLSTM as the schema representations $\pmb x_i^{(s)}$.
$X$ is the set of all the obtained representations from $Q\cup\mathcal{S}$: $X = (\pmb x_1^{(q)}, \dots, \pmb x_{\lvert Q\rvert}^{(q)}, \pmb x_1^{(s)}, \dots, \pmb x_{\lvert \mathcal{S}\rvert}^{(s)}).$ Along with the relational embeddings $\pmb r^k, \pmb r^v$ specified by $\mathcal{G}$, $X$ is passed into the main transformer module.
\vspace{-1.5em}
\subsection{Schema Linking}
\vspace{-.5em}
The goal of schema linking is to identify the implicit relations between $Q$ and $\mathcal{S}$.
The relations are defined by whether there exist column/table references in the question to the corresponding schema columns/tables, given certain heuristics.
Following \citet{wang2019rat}, possible relations for each $(i, j)$ where $x_i \in Q, x_j \in \mathcal{S}$ (or vice versa) can be {\tt ExactMatch}, {\tt PartialMatch}, or {\tt NoMatch}, which are based on name-based linking.
Depending on the type of $x_i$ and $x_j$, the above three relations are further expanded to four types: {\tt Question-Column}, {\tt Question-Table}, {\tt Column-Question}, or {\tt Table-Question}.
We also use the value-based linking from \citet{wang2019rat} and \citet{guo2019towards} to augment the {\tt ExactMatch} relation by database content and external knowledge.
%Furthermore, we add a couple of heuristics to address the low precision issue we observed in the original schema linking method.
\vspace{-.5em}
\subsection{Decoder}
\vspace{-.5em}
For our decoder (as the post-transformer module) $f_o$, we employ a transition-based abstract syntax decoder following \citet{yin2018tranx}. 
It requires a transition system to converts between the surface SQL and a AST-tree constructing action sequences, and can ensure grammarticality of generation. The neural model then predicts the action sequences.
There are three types of actions to generate the target SQL $T$, including (i) {\tt ApplyRule} which applies a production rule to the last generated node; (ii) {\tt Reduce} which completes a leaf node; (iii) {\tt SelectColumn} which chooses a column from the schema.
For our transition system, each column is attached with their corresponding table so that the tables in the target SQL $T$ can be directly inferred from the predicted columns.
As a result, action {\tt SelectTable} can be omitted from the generation.
Formally, the generation process can be formulated as $\Pr(T|\mathcal{Y}) = \prod_t \Pr(a_t|a_{<t}, \mathcal{Y})$ where $\mathcal{Y}$ is the outputs of the last layer of the relational transformers.
We use a parent-feeding LSTM as the decoder.
The LSTM state is updated as $\pmb m_t, \pmb h_t=f_{\text{LSTM}}([\pmb a_{t-1} \Vert \pmb z_{t-1} \Vert \pmb h_{p_t} \Vert \pmb a_{p_t} \Vert \pmb n_{p_t}], \pmb m_{t-1}, \pmb h_{t-1})$, where $\pmb m_t$ is the LSTM cell state, $\pmb h_t$ is the LSTM output at step $t$, $\pmb a_{t-1}$ is the action embedding of the previous step, $\pmb z_{t-1}$ is the context feature computed using multi-head attention on $\pmb h_{t-1}$ over $\mathcal{Y}$, $p_t$ is the step corresponding to the parent AST node of the current node, and $\pmb n$ is the node type embedding.
For {\tt ApplyRule[R]}, we compute $\Pr(a_t=\texttt{ApplyRule[R]}|a_{<t}, y)=\text{softmax}_R(g(\pmb z_t))$ where $g(\cdot)$ is a $2$-layer MLP.
For {\tt SelectColumn}, we use the memory augmented pointer net \citet{guo2019towards}.
%We refer the reader to \citet{guo2019towards} for details.
\vspace{-.5em}
\subsection{Regularization}
\vspace{-.5em}
Besides using dropout \cite{srivastava2014dropout} employed on $X$ and $\pmb z_t$ to help regularize the model, we further apply uniform label smoothing \cite{szegedy2016rethinking} on the objective of predicting {\tt SelectColumn}.
Formally, the cross entropy for a ground-truth column $c^*$ we optimize becomes: $(1-\epsilon) * \log p(c^*) + {\epsilon}/{K} * {\textstyle \sum}_c\log p(c)$, where $K$ is the number of columns in the schema, $\epsilon$ is the weight of the label smoothing term, and $p(\cdot) \triangleq \Pr(a_t=\texttt{SelectColumn}[\cdot]|a_{<t}, y)$.
\vspace{-.5em}
\subsection{Experiment Configuration}
\vspace{-.5em}
We choose RoBERTa \cite{liu2019roberta} as the pre-trained language models.
A sequence of 24 relation-aware transformer layers are stacked on top of $f_e$.
% For \mname, we choose the initial weight shrinkage parameter $\alpha$ as $15$.
The Adam optimizer \cite{kingma2014adam} with the default hyperparameters is used to train the model with an initial learning rate $\eta$ of $4 \times 10^{-4}$.
$\eta$ is annealed to $0$ with $4 \times 10^{-4}(1-steps/max\_steps)^{0.5}$.
A separate learning rate is used to fine-tune the RoBERTa by multiplying $\eta$ a factor of $8 \times 10^{-3}$.
The BiLSTM to encode the schema representations has hidden size $128$ per direction.
For each transformer layer, $d_x=d_z=256$, $H=8$ and the inner layer dimension of the position-wise MLP is $1024$.
For the decoder, we use action embeddings of size $128$, node type embeddings of size of $64$, and LSTM hidden state of size $512$.
We apply dropout rate of $0.6$ on the input to the relational transformers $X$ and the context representation $\pmb{z}_t$.
The weight of the label smoothing term is set to be $0.2$.
We use a batch size of $16$ and train $60$ epochs (around $25,000$ steps).
During inference, beam search is used with beam size as $5$.
%% \vspace{-.5em}
%% \subsection{Hyperparameter Tuning}
%% \vspace{-.5em}
Most of the hyperparameters are chosen following \citet{wang2019rat}.
We only tune the learning rate ($4 \times 10^{-4}$ to $8 \times 10^{-4}$ with step size $1 \times 10^{-4}$), dropout ($0.3$, $0.4$, $0.5$, $0.6$), the weight of the label smoothing $\epsilon$ ($0.0$, $0.1$, $0.2$) by grid search.
The average runtime is around 30 hours and the number of parameters is around 380 millions.
    
\vspace{-.5em}
\section{Implementation Details for Logical Reading Comprehension}\label{sec:mrc}
\vspace{-.5em}
We build on the code\footnote{https://github.com/yuweihao/reclor} by \citet{yu2020reclor} and use it for evaluation. For each example, the encoder embeds the input context, question and options which are then passed to the linear layer for classification. The exact input format to the encoder is ``$\langle s \rangle$ Context $\langle /s \rangle \langle /s \rangle$ Question  $ || $ Option $\langle pad \rangle \dots$'', where ``$||$'' denotes concatenation. The linear layer uses the embedding of the first token $\langle s \rangle$ for classification.
\vspace{-.5em}
\subsection{Experimental Configuration}
\vspace{-.5em}
RoBERT is chosen as the pre-trained model, and we stack $4$ transformer layers on top. The Adam optimizer \cite{kingma2014adam} with  $\epsilon = 10^{-6}$ and betas of $(0.9, 0.98)$ is used. The learning rate to finetune RoBERTa is $1 \times 10^{-5}$ while the learning rate for the additional transformer layers is $3 \times 10^{-4}$.
%with DT-Fixup and the standard case is $3 \times 10^{-4}$ and $1 \times 10^{-4}$ respectively. 
For all models in our ablation study, the learning rate for the additional transformer layers is $1 \times 10^{-4}$. The learning rate is annealed linearly to $0$ with weight decay of $0.01$. We use a batch size of $24$ and fine-tune for $12$ epochs. % The max input length is $256$.
For each transformer layer, $d_x=d_z=1024$, $H=8$ and the inner layer dimension of the position-wise MLP is $2048$. We use dropout rate of $0.4$ on the input to the additional transformer layers and $0.1$ for the linear layer.
%% We choose RoBERTa \cite{liu2019roberta} as the pre-trained language model and use the Transformers library of Hugging Face\footnote{https://github.com/huggingface/transformers}. A sequence of 16 transformer layers are stacked on top of it. The Adam optimizer \cite{kingma2014adam} with epsilon value of $1 \times 10^{-6}$ and beta values of $(0.9, 0.98)$ is used to train the model. The learning rate to finetune RoBERTa is $1 \times 10^{-5}$ where as the learning rate to train the transformer layers is $1 \times 10^{-4}$. For learning rate is annealed linearly to $0$. Weight decay of $0.01$ is also applied. We use a batch size of $24$ and fine-tune for $12$ epochs. The maximum input sequence length is $256$. 
%% For each transformer layer, $d_x=d_z=1024$, $H=8$ and the inner layer dimension of the position-wise MLP is $2048$. We apply dropout rate of $0.4$ to the transformer layers and $0.1$ to the linear layer.
We follow the hyperparameters used in \citet{yu2020reclor} for the pretrained language model. %and the linear layer. 
For the additional transformer layers, we only tune the dropout values $(0.3, 0.4, 0.5, 0.6)$. The average runtime is around 6 hours and the number of parameters is around 39 millions.

\end{document}